\documentclass[10pt,twocolumn,letterpaper]{article}
\pdfoutput=1

\usepackage{cvpr}
\usepackage{times}
\usepackage{epsfig}
\usepackage{graphicx}
\usepackage{amsmath}
\usepackage{amssymb}
\usepackage{lipsum}
\usepackage{float}
\usepackage[normalem]{ulem}

\usepackage{color}
\usepackage{amsmath}
\usepackage{amssymb}
\usepackage{multirow, varwidth}
\usepackage{verbatim}
\makeatletter
\newif\if@restonecol
\makeatother

\usepackage[ruled,vlined]{algorithm2e}
\usepackage{xr}
\usepackage[amsmath,thmmarks]{ntheorem}

\theorembodyfont{\normalfont}
\theoremseparator{.}
\theoremstyle{nonumberplain}

\newcommand{\bm}[1]{\boldsymbol{#1}}
\newcommand{\argmax}{\operatornamewithlimits{argmax}}

\DeclareRobustCommand\onedot{\futurelet\@let@token\@onedot}
\def\onedot{.\@\xspace}

\def\eg{\emph{e.g}\onedot} 
\def\ie{\emph{i.e}\onedot}

\def\etal{\emph{et al}\onedot}

\newcommand{\x}{\mathbf{x}}

\usepackage{placeins}
\usepackage{morefloats}
\usepackage{dblfloatfix}
\usepackage{booktabs}

\usepackage{color}

\usepackage[pagebackref=true,breaklinks=true,letterpaper=true,colorlinks,bookmarks=false]{hyperref}

\cvprfinalcopy %

\def\cvprPaperID{1165} %

\ifcvprfinal\fi
\begin{document}

\title{End-to-end Concept Word Detection \\for Video Captioning, Retrieval, and Question Answering} %

\author{Youngjae Yu \hspace{9pt} Hyungjin Ko \hspace{9pt}  Jongwook Choi \hspace{9pt} Gunhee Kim \\
Seoul National University, Seoul, Korea\\
{\tt\small \{yj.yu, hj.ko\}@vision.snu.ac.kr, \{wookayin, gunhee\}@snu.ac.kr } \\
{\small \url{http://vision.snu.ac.kr/projects/lsmdc-2016}}
}

\maketitle\thispagestyle{empty}

\begin{abstract}
We propose a high-level concept word detector that can be integrated with any video-to-language models.
It takes a video as input and generates a list of concept words as useful semantic priors for language generation models.
The proposed word detector has two important properties. First, it does not require any external knowledge sources for training.
Second, the proposed word detector is trainable in an end-to-end manner jointly with any video-to-language models.
To effectively exploit the detected words,
we also develop a semantic attention mechanism that selectively focuses on the detected concept words and fuse them with the word encoding and decoding in the language model.
In order to demonstrate that the proposed approach indeed improves the performance of multiple video-to-language tasks,
we participate in all the four tasks of LSMDC 2016~\cite{rohrbach-ijcv-2017}.
Our approach has won three of them, including \textit{fill-in-the-blank}, \textit{multiple-choice test}, and \textit{movie retrieval}.
\end{abstract}

\vspace{-5pt}
\section{Introduction}
\label{sec:introduction}

\begin{figure}[t]
\centering
\includegraphics[trim=0.2cm 0.2cm 0cm 0.1cm,clip,width=0.47\textwidth]{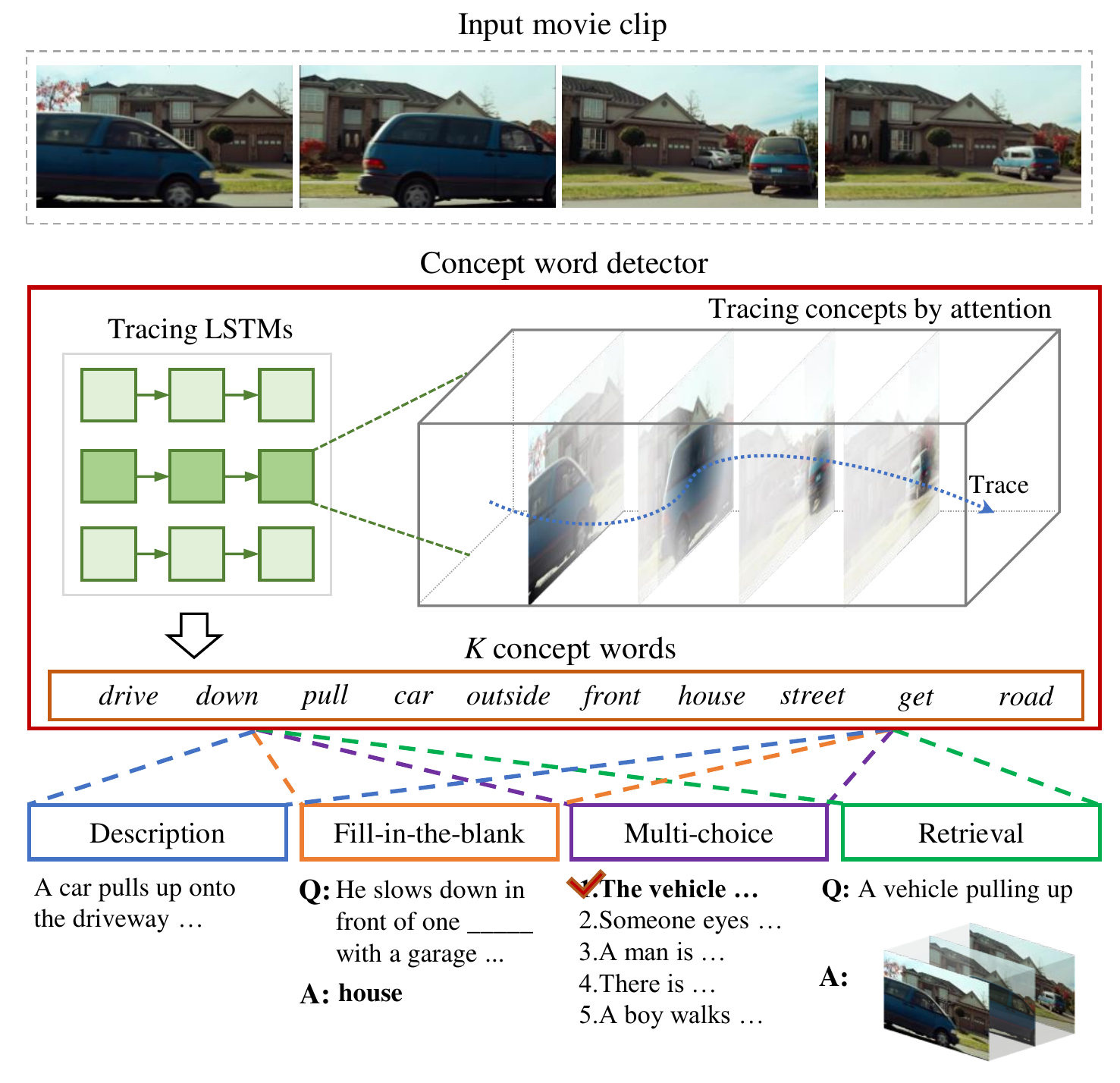}
\caption{The intuition of the proposed concept word detector. Given a video clip,
    a set of tracing LSTMs extract multiple concept words that consistently appear across frame regions.
    We then employ semantic attention to combine the detected concepts with text encoding/decoding
    for several video-to-language tasks of LSMDC 2016, such as captioning, retrieval, and question answering.
}
\vspace{-5pt}
\label{fig:keyidea}
\end{figure}

Video-to-language tasks, including video captioning~\cite{jeff-cvpr-2015,guadarrama-iccv-2013,rohrbach-gcpr-2015,venugopalan-iccv-2015,xu-aaai-2015,yu-cvpr-2016} and video question answering (QA)~\cite{tapaswi-cvpr-2016}, are recent emerging challenges in computer vision research.
This set of problems is interesting as one of frontiers in artificial intelligence; beyond that, it can also potentiate multiple practical applications, such as retrieving video content by users' free-form queries or  helping visually impaired people understand the visual content.
Recently, a number of large-scale datasets have been introduced as a common ground for researchers to promote the progress of video-to-language research
(\eg \cite{chen-acl-2011,rohrbach-gcpr-2014, rohrbach-ijcv-2017,tapaswi-cvpr-2016}).

The objective of this work is to propose a \emph{concept word detector}, as shown in Fig.\ref{fig:keyidea},
which takes a training set of videos and associated sentences as input, and generates a list of high-level concept words per video as useful semantic priors for a variety of video-to-language tasks, including video captioning, retrieval, and question answering.
We design our word detector to have the following two characteristics, to be easily integrated with any video-to-language models.
First, it does not require any external knowledge sources for training.
Instead, our detector learns the correlation between words in the captions and video regions from the whole training data.
To this end, we use a continuous soft attention mechanism that traces consistent visual information across frames and associates them with concept words from captions.
Second, the word detector is trainable in an end-to-end manner jointly with any video-to-language models.
The loss function for learning the word detector can be plugged as an auxiliary term into the model's overall cost function; as a result, we can reduce efforts to separately collect training examples and learn both models.

We also develop language model components to to effectively exploit the detected words. 
Inspired by \textit{semantic attention} in image captioning research~\cite{quanzeng-cvpr-2016},
we develop an attention mechanism that selectively focuses on the detected concept words and fuse them with word encoding and decoding in the language model.
That is, the detected concept words are combined with input words to better represent the hidden states of encoders, and with output words to generate more accurate word prediction.

In order to demonstrate that the proposed word detector and attention mechanism indeed improve the performance of multiple video-to-language tasks,
we participate in four tasks of LSMDC 2016 (\textit{Large Scale Movie Description Challenge})~\cite{rohrbach-ijcv-2017},
which is  one of the most active and successful benchmarks that advance the progress of video-to-language research.
The challenges include \textit{movie description} and \textit{multiple-choice test} as video captioning, %
\textit{fill-in-the-blank} as video question answering,
and \textit{movie retrieval} as video retrieval.
Following the public evaluation protocol of LSMDC 2016, our approach achieves the best accuracies in the three tasks  (\textit{fill-in-the-blank}, \textit{multiple-choice test}, and \textit{movie retrieval}),
and comparable performance in the other task (\textit{movie description}).

\subsection{Related Work}
\label{sec:related_work}

Our work can be uniquely positioned in the context of two recent research directions in image/video captioning. %

\textbf{Image/Video Captioning with Word Detection}.
Image and video captioning has been actively studied in recent vision and language research, including \cite{das-cvpr-2013,jeff-cvpr-2015,guadarrama-iccv-2013,rohrbach-gcpr-2015,rohrbach-iccv-2013,venugopalan-iccv-2015,venugopalan-hlt-2015}, to name a few.
Among them, there have been several attempts to detect a set of concept words or attributes from visual input to boost up the captioning performance.
In image captioning research,
Fang \etal~\cite{fang-cvpr-2015} exploit a multiple instance learning (MIL) approach to train visual detectors that identify a set of words with bounding boxed regions of the image.
Based on the detected words, they retrieve and re-rank the best caption sentence for the image.
Wu \etal~\cite{Wu:2016:External} use a CNN to learn a mapping between an image and semantic attributes.
They then exploit the mapping as an input to the captioning decoder.
They also extend the framework to explicitly leverage external knowledge base such as DBpedia for question answering tasks.
Venugopalan \etal~\cite{Venugopalan:2016:NOC} generate description with novel words beyond the ones in the training set,
by leveraging external sources, including object recognition datasets like ImageNet and external text corpus like Wikipedia.
You \etal\cite{quanzeng-cvpr-2016} also exploit weak labels and tags on Internet images to train additional parametric visual classifiers for image captioning.

In the video domain, it is more ambiguous to learn the relation between descriptive words and visual patterns.
There have been only few work in video captioning. %
Rohrbach \etal\cite{rohrbach-gcpr-2015} propose a two-step approach for video captioning on the LSMDC dataset.
They first extract verbs, objects, and places from movie description, and separately train SVM-based classifiers for each group.
They then learn the LSTM decoder that generates text description based on the responses of these visual classifiers.

While almost all previous captioning methods exploit external classifiers for concept or attribute detection,
the novelty of our work lies in that we use only captioning training data with no external sources to learn the word detector,
and propose an end-to-end design for learning both word detection and caption generation simultaneously.
Moreover, compared to video captioning work of \cite{rohrbach-gcpr-2015} where only \textit{movie description} of LSMDC is addressed,
this work is more comprehensive in that we validate the usefulness of our method for all the four tasks of LSMDC.

\textbf{Attention for Captioning}.
Attention mechanism has been successfully applied to caption generation.
One of the earliest works is \cite{xu-icml-2015} that dynamically focuses on different image regions to produce an output word sequence.
Later this soft attention has been extended as temporal attention over video frames \cite{yao-iccv-2015,yu-cvpr-2016} for video captioning.

Beyond the attention on spatial or temporal structure of visual input,
recently You \etal\cite{quanzeng-cvpr-2016} propose an attention on attribute words for image captioning.
That is, the method enumerates a set of important object labels in the image, and then dynamically switch attention among these concept labels.
Although our approach also exploits the idea of semantic attention, it bears two key differences.
First, we extend the semantic attention to video domains for the first time,
not only for video captioning but also for retrieval and question answering tasks.
Second, the approach of \cite{quanzeng-cvpr-2016} relies on the classifiers that are separately learned from external datasets,
whereas our approach is learnable end-to-end with only training data of captioning.
It significantly reduces  efforts to prepare for additional multi-label classifiers.

\subsection{Contributions}
\label{sec:contributions}

We summarize the contributions of this work as follows.

(1)  We propose a  novel end-to-end learning approach for detecting a list of concept words and attend on them to enhance the performance of multiple video-to-language tasks.
The proposed concept word detection and attention model can be plugged into any models of video captioning, retrieval, and question answering.
Our technical novelties can be seen from two recent trends of image/video captioning research.
First, %
our work is a first end-to-end trainable model not only for concept word detection but also for language generation.
Second, our work is a first semantic attention model for video-to-language tasks. %

(2) To validate the applicability of the proposed approach, we participate in all the four tasks of LSMDC 2016.
Our models have won three of them, including \textit{fill-in-the-blank}, \textit{multiple-choice test}, and \textit{movie retrieval}.
We also attain comparable performance for \textit{movie description}.

\section{Detection of Concept Words from Videos}
\label{sec:approach}

We first explain the pre-processing steps for representation of words and video frames.
Then, we explain how we detect concept words for a given video.

\subsection{Preprocessing}
\label{sec:preproc}

\textbf{Dictionary and Word Embedding}.
We define a vocabulary dictionary $\mathcal V$ by collecting the words that occur more than three times in the dataset.
The dictionary size is $|\mathcal V| = 12\,486$, from which our models sequentially select words as output.
We train the word2vec skip-gram embedding~\cite{Tomas-nips-2013} to obtain the word embedding matrix
$\mathbf E \in \mathbb R^{d \times |\mathcal V|}$ where $d$ is the word embedding dimension and $V$ is the dictionary size.
We set $d=300$ in our implementation.

\textbf{Video Representation}.
We first equidistantly sample one per ten frames from a video, to reduce the frame redundancy while minimizing loss of information.
We denote the number of video frames by $N$.
We limit the maximum number of frames to be $N_{max}=40$;
if a video is too long, we use a wider interval for uniform sampling.

We employ a convolutional neural network (CNN) to encode video input.
Specifically, we extract %
the feature map of each frame from the res5c layer
(\ie $\mathbb R^{7 \times 7 \times 2,048}$)
of ResNet \cite{he-arxiv-2015} pretrained on ImageNet dataset~\cite{imagenet-ijcv-2015},
and then apply a $2 \times 2$ max-pooling followed by a $3 \times 3$ convolution
to reduce dimension to $\mathbb R^{4 \times 4 \times 500}$.
Reducing the number of spatial grid regions to $4 \times 4$ helps the concept word detector
get trained much faster, while not hurting detection performance significantly. %
We denote resulting visual features of frames by $\{\mathbf v_n \}_{n=1}^N$.
Throughout this paper, we use $n$ for denoting video frame index.

\begin{figure*}%
\centering
\includegraphics[width=0.95\textwidth,trim=0cm 0cm 0.8cm 0cm]{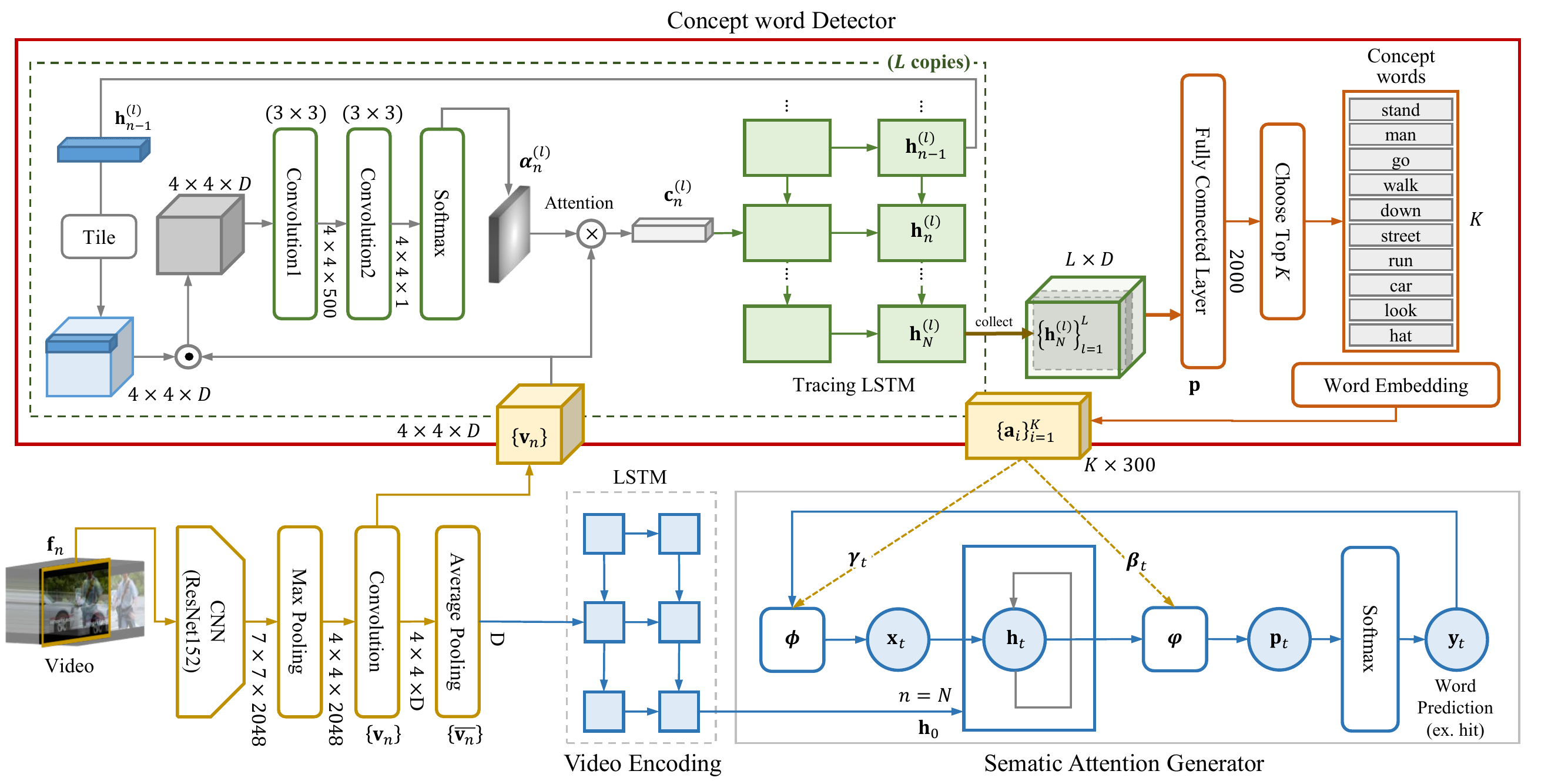}
\caption{
    The architecture of the concept word detection in a top red box (section \ref{subsec:model_attributeword}),
    and our video description model in bottom, which uses semantic attention on the detected concept words (section \ref{subsec:model_description}).
}
\label{fig:model_vidcap}
\vspace{-5pt}
\end{figure*}

\subsection{An Attention Model for Concept Detection}
\label{subsec:model_attributeword}

\textbf{Concept Words and Traces.}
We propose the \emph{concept word detector} using LSTM networks with soft attention mechanism.
Its structure is shown in the red box of Fig.\ref{fig:model_vidcap}.
Its goal is, for a given video, to discover a list of \emph{concept words} that consistently appear across frame regions.
The detected concept words are used as additional references for video captioning models (section \ref{subsec:model_description}),
which generates output sentence by selectively attending on those words.

We first define a set of candidate words with a size of $V$ from all training captions.
Among them, we discover $K$ concept words per video. %
We set $V=2,000$ and $K=10$. %
We first apply the automatic POS tagging of NLTK \cite{nltk}, to extract nouns, verbs and adjectives from all training caption sentences~\cite{fang-cvpr-2015}.
We then compute the frequencies of those words in a training set, and select the $V$ most common words as concept word candidates.

Since we do not have groundtruth bounding boxes for concept words in the videos, we cannot train individual concept detectors in a standard supervised setting.
Our idea is to adopt a soft attention mechanism to infer words by tracking regions that are spatially consistent.  %
To this end,
we employ a set of \emph{tracing LSTMs}, each of which takes care of a single spatially-consistent meaning
being tracked over time, what we call \emph{trace}.
That is, we keep track of spatial attention over video frames using an LSTM,
so that spatial attentions in adjacent frames resemble the spatial consistency
of a single concept (\eg a moving object, or an action in video clips; see Fig.\ref{fig:keyidea}).
We use a total of $L$ tracing LSTMs to capture out $L$ traces (or concepts),
where $L$ is the number of spatial regions in the visual feature (\ie $L = 4 \times 4 = 16$ for $\mathbf v \in \mathbb{R}^{4 \times 4 \times D}$).
Fusing these $L$ concepts together, we finally discover $K$ concept words,
as will be described next.

\textbf{Computation of Spatial Attention.}
\newcommand{\bfalpha}{\boldsymbol{\alpha}}
For each trace $l$, we maintain %
spatial attention weights $\bfalpha^{(l)}_n \in \mathbb R^{4\times4}$,
indicating where to attend on $(4\times4)$ spatial grid locations of $\mathbf v_n$,
through video frames $n=1\ldots N$.
The initial attention weight $\bfalpha^{(l)}_0$ at $n = 0$ is initialized with an one-hot matrix,
for each of $L$ grid locations.
We compute the hidden states $\mathbf h^{(l)}_n \in \mathbb R^{500}$ of the LSTM
through $n = 1\ldots N$ by:
\begin{align}
    \label{eq:concept_lstm}
    \mathbf c^{(l)}_n &=
        \bfalpha^{(l)}_n \otimes \mathbf v_n \\
    \label{eq:concept_lstm2}
    \mathbf h^{(l)}_n &=
        \mbox{LSTM}(\mathbf c^{(l)}_n , \mathbf h^{(l)}_{n-1}).
\end{align}%
where $A\otimes B = \sum_{j,k} A_{(j,k)} \cdot B_{(j,k,:)}$.
The input to LSTMs is the context vector $\mathbf c^{(l)}_n \in \mathbb R^{500}$,
which is obtained by applying spatial attention $\bfalpha^{(l)}_n$ to the visual feature $\mathbf v_n$.
Note that the parameters of $L$ LSTMs are shared.

The attention weight vector $\bfalpha^{(l)}_n \in \mathbb{R}^{4 \times 4}$
at time step $n$ is updated as follows:
\begin{align}
\label{eq:concept_attention}
    \mathbf e^{(l)}_n(j, k) &= \mathbf v_n(j,k) \odot \mathbf h^{(l)}_{n-1}, \\
    \bfalpha^{(l)}_n        &= \mathrm{softmax} \left( \mbox{Conv}( \mathbf e^{(l)}_n ) \right),
\end{align}%
where $\odot$ is elementwise product, %
and $\mbox{Conv}(\cdot)$ denotes two convolution operations before the softmax layer in Fig.\ref{fig:model_vidcap}.
Note that $\bfalpha^{(l)}_n$ in Eq.(\ref{eq:concept_attention}) is computed from the previous hidden state $\mathbf h^{(l)}_{n-1}$ of the LSTM.

The spatial attention $\bfalpha^{(l)}_n$ measures
how each spatial grid location of visual features is related to the concept being tracked through tracing LSTMs.
By repeating these two steps of Eq.(\ref{eq:concept_lstm})--(\ref{eq:concept_attention}) from $n=1$ to $N$,
our model can continuously find important and temporally consistent meanings over time,
that are closely related to a part of video,
rather than focusing on each video frame individually.

Finally, we predict the concept confidence vector $\mathbf {p}$:
\begin{align}
    \label{eq:concept_confidence_p}
    \mathbf p &= \sigma \left(\mathbf W_p \left[ \mathbf h^{(1)}_N; \cdots; \mathbf h^{(L)}_N \right] + \mathbf b_p \right)
      \in \mathbb R^{V},
\end{align}
that is, we first concatenate the hidden states $\{ \mathbf h_N^{(l)} \}_{l=1}^L$ at the last time step of all tracing LSTMs,
apply a linear transform parameterized by $\mathbf W_p \in \mathbb R^{V \times (500L)}$ and $\mathbf b_p \in \mathbb R^{V}$,
and apply the elementwise sigmoid activation $\sigma$.

\medskip
\textbf{Training and Inference.}
For training, we obtain a reference concept confidence vector $\mathbf p^* \in \mathbb R^{V}$
whose element $p^*_i$ is 1 if the corresponding word exists in the groundtruth caption; otherwise, 0.
We minimize the following sigmoid cross-entropy cost $\mathcal{L}_{con}$,
which is often used for multi-label classification \cite{wu-cvpr-2016} where each class is independent and not mutually exclusive:
\begin{align}
    \label{eq:concept_cost}
    \mathcal L_{con} = - \frac{1}{V} \sum_{i=1}^{V} \left[ p^*_i \log(p_i) + (1 - p^*_i) \log(1 - p_i) \right] .
\end{align}%
Strictly speaking, since we apply an end-to-end learning approach,
the cost of Eq.(\ref{eq:concept_cost}) is used as an auxiliary term for the overall cost function, which will be discussed in section \ref{sec:v2l_models}.

For inference, %
we compute $\mathbf p$ for a given query video, and find top $K$ words from the score $\mathbf p$ (\ie $\argmax_{1:K} \mathbf p$).
Finally, we represent these $K$ concept words by their word embedding $\{ \mathbf a_i \}_{i=1}^K$.

\section{Video-to-Language Models}
\label{sec:v2l_models}

We design a different base model for each of LSMDC tasks, while they share the concept word detector and the semantic attention mechanism.
That is, we aim to validate that the proposed concept word detection is useful to a wide range of video-to-language models.
For base models, we take advantage of state-of-the-art techniques, for which we do not argue as our contribution.
We refer to our video-to-language models leveraging the concept word detector
as \emph{CT-SAN} (\emph{Concept-Tracing Semantic Attention Network}).

For better understanding of our models, we outline the four LSMDC tasks as follows:
(i) \textit{Movie description}: generating a single descriptive sentence for a given movie clip,
(ii) \textit{Fill-in-the-blank}: given a video and a sentence with a single blank, finding a suitable word for the blank from the whole vocabulary set,
(iii) \textit{Multiple-choice test}: given a video query and five descriptive sentences, choosing the correct one out of them, and
(iv) \textit{Movie retrieval}: ranking 1,000 movie clips for a given natural language query.

We defer more model details to the supplementary file.
Especially, we skip the description of multiple-choice and movie retrieval models in Figure \ref{fig:models_fib_mc_ret}(b)--(c),
which can be found in the supplementary file.

\subsection{A Model for Description}
\label{subsec:model_description}

Fig.\ref{fig:model_vidcap} shows the proposed video captioning model.
It takes video features $\{ \mathbf v_n \}_{n=1}^N$ and the detected concept words $\{ \mathbf a_i \}_{i=1}^K$ as input,
and produces a word sequence as output $\{ \mathbf y_t \}_{t=1}^T$.
The model comprises video encoding and caption decoding LSTMs, and two semantic attention models.
The two LSTM networks have two layers in depth, with layer normalization~\cite{Jimmy-arxiv-2016} and dropout~\cite{srivastava-jmlr-2014} with a rate of 0.2.

\textbf{Video Encoder.}
The \textit{video encoding LSTM} encodes a video into a sequence of hidden states $\{ \mathbf s_n \}_{n=1}^N \in \mathbb R^{D}$.
\begin{align}
    \mathbf s_n &=   \mbox{LSTM} (\overline{\mathbf v_n} , \mathbf s_{n-1})
\end{align}%
where $\overline{\mathbf v_n} \in \mathbb R^{D}$ is obtained by $(4,4)$-average-pooling $\mathbf v_n$.

\textbf{Caption Decoder.}
The \textit{caption decoding LSTM} is a normal LSTM network as follows:
\begin{align}
\label{eq:decoding_lstm}
\mathbf h_t =  \mbox{LSTM} (\mathbf x_t , \mathbf h_{t-1} ),
\end{align}
where the input $\mathbf x_t$ is an intermediate representation of $t$-th word input with semantic attention applied, as will be described below.
We initialize the hidden state at $t=0$ by the last hidden state of the video encoder: $\mathbf h_{0} = \mathbf s_{N} \in \mathbb R^{D}$.

\textbf{Semantic Attention.}
Based on \cite{quanzeng-cvpr-2016}, our model in Fig.\ref{fig:model_vidcap} uses the semantic attention in two different parts,
which are called as \textit{input} and \textit{output} semantic attention, respectively.

The \emph{input semantic attention} $\phi$ computes an attention weight $\gamma_{t,i}$,
which is assigned to each predicted concept word $\mathbf a_i$.
It helps the caption decoding LSTM focus on different concept words dynamically at each step $t$.

The attention weight $\gamma_{t,i} \in \mathbb R^K$ and input vector $\mathbf x_t \in \mathbb R^D$ to the LSTM are obtained by
\begin{align}
\label{eq:attention_input_wgt}
    \gamma_{t,i}    &\propto \exp( {(\mathbf E \mathbf y_{t-1})}^\top \mathbf W_{\gamma} \mathbf a_i) , \\
    \mathbf x_t             &= \phi (\mathbf y_{t-1}, \{ \mathbf a_i \} ) \nonumber \\
\label{eq:attention_input_vector}
    						&= \mathbf W_x (\mathbf{E y}_{t-1} + \mathrm{diag}(\mathbf w_{x,a}) \sum_i \gamma_{t,i} \mathbf a_i).
\end{align}%
We multiply a previous word $\mathbf y_{t-1} \in \mathbb R^{|\mathcal V|}$ by the word embedding matrix $\mathbf E$ to be $d$-dimensional.
The parameters to learn include $\mathbf W_{\gamma} \in \mathbb R^{d \times d}$, $\mathbf W_{x} \in \mathbb R^{D \times d}$  and $\mathbf w_{x,a} \in \mathbb R^d$.

The \emph{output semantic attention} $\varphi$ guides how to dynamically weight the concept words $ \{\mathbf a_i\}$ when generating an output word $\mathbf y_{t}$ at each step.
We use $\mathbf h_t$, the hidden state of decoding LSTM at $t$ as an input to the output attention function $\varphi$.
We then compute $\mathbf p_t \in \mathbb R^D$ by attending the concept words set $\{\mathbf a_i \}$ with the weight $\beta_{t,i}$:
\begin{align}
\label{eq:attention_out_wgt}
    \beta_{t,i}     &\propto \exp(\mathbf h_t^\top \mathbf W_{\beta} \sigma(\mathbf a_i)) , \\
\label{eq:attention_out_vec}
    \mathbf p_t     &= \varphi (\mathbf h_t, \{ \mathbf a_i \} ) \nonumber \\
                    &= \mathbf h_t + \mathrm{diag} (\mathbf w_{h,a} ) \sum_i \beta_{t,i} \mathbf W_{\beta} \sigma(\mathbf a_i),
\end{align}
where $\sigma$ is the hyperbolic tangent,
and parameters include $\mathbf w_{h,a} \in \mathbb R^{D}$
and $\mathbf  W_\beta \in \mathbb R^{D \times d}$.

Finally, the probability of output word is obtained as
\begin{align}
    p(\mathbf y_t \mid \mathbf y_{1:t-1})    &= \mathrm{softmax} ( \mathbf W_y \mathbf p_{t} + \mathbf b_y),
\end{align}%
where $\mathbf W_y \in \mathbb R^{|\mathcal V| \times D}$ and $\mathbf b_y \in \mathbb{R}^{|\mathcal V|}$.
This procedure loops until $ \mathbf y_t $ corresponds to the \texttt{<EOS>} token.

\textbf{Training.}
To learn the parameters of the model, we define a loss function as the total negative log-likelihood of all the words,
with regularization terms on attention weights
$\{ \mathbf \alpha_{t,i} \}$, $\{ \mathbf \beta_{t,i} \}$, and $\{ \mathbf \gamma_{t,i} \}$ \cite{quanzeng-cvpr-2016},
as well as the loss $\mathcal{L}_{con}$ for concept discovery (Eq.\ref{eq:concept_cost}):
\begin{align}
    \label{eq:loss_function}
    \mathcal{L} = - \sum_{t} \log p(\mathbf y_t) + \lambda_1 (g(\mathbf \beta) + g(\mathbf \gamma)) + \lambda_2 \mathcal{L}_{con}
\end{align}%
where $\lambda_1, \lambda_2$ are hyperparameters and $g$ is a regularization function with setting to $p=2, q=0.5$  as
\begin{align}
\label{eq:regularizer}
	g(\mathbf \alpha)	&= \|\mathbf \alpha\|_{1,p} + \|\mathbf \alpha^\top \|_{1,q} \\
						&= \left [ \sum_i \left [ \sum_t \mathbf \alpha_{t,i} \right ]^p \right ]^{1/p}
+ \left [ \sum_t \left [ \sum_i \mathbf \alpha_{t,i} \right ]^q \right ]^{1/q}. \nonumber
\end{align}%

For the rest of models, we transfer the parameters of the concept word detector
trained with the description model, %
and allow the parameters being fine-tuned.

\begin{figure*}%
\centering
\includegraphics[trim=0cm 0.05cm 0cm 0cm,clip,width=0.99\textwidth]{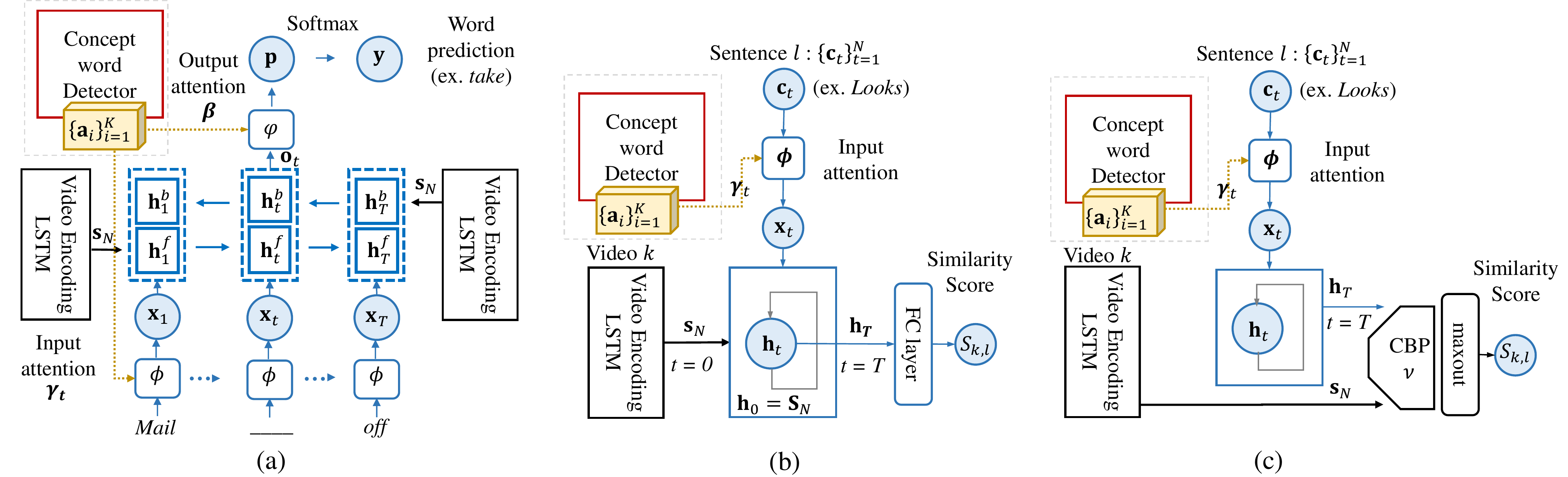}
\vspace{5pt}
\caption{The model architectures for
    (a) fill-in-the-blank (section \ref{subsec:model_blank}),
    (b) multiple-choice, %
    and (c) movie retrieval task. %
The description of models for (b)--(c) can be found in the supplementary file.
Each model takes advantage of the concept word detector in Fig.\ref{fig:model_vidcap},
and semantic attention for the sake of its objective.
}
\label{fig:models_fib_mc_ret}
\vspace{-5pt}
\end{figure*}

\subsection{A Model for Fill-in-the-Blank}
\label{subsec:model_blank}

Fig.\ref{fig:models_fib_mc_ret}(a) illustrates the proposed model for the fill-in-the-blank task.
It is based on a \textit{bidirectional LSTM network} (BLSTM)~\cite{Schuster-ieee-1997,hochreiter-ieee-1997},
which is useful in predicting a blank word from an imperfect sentence, since it considers the sequence in both forward and backward directions.
Our key idea is to employ the semantic attention mechanism on both input and output of the BLSTM,
to strengthen the meaning of input and output words with the detected concept words. %

The model takes word representation $\{ \mathbf c_t \}_{t=1}^T$ and concept words $\{ \mathbf a_i \}_{i=1}^K$ as input.
Each $\mathbf c_t \in \mathbb R^d$ is obtained by multiplying the one-hot word vector by an embedding matrix $\mathbf E$.
Suppose that the $t$-th text input is a blank for which we use a special token \texttt{<blank>}.
We add the word prediction module only on top of the $t$-th step of the BLSTM.

\textbf{BLSTM.}
The input video is represented by the \textit{video encoding LSTM} in Figure \ref{fig:model_vidcap}.
The hidden state of the final video frame $\mathbf s_N$ is used to
initialize the hidden states of the BLSTM:
$\mathbf{h}^b_{T+1} = \mathbf{h}^f_0 = \mathbf s_N$, where
$\{ \mathbf h^f_t \}_{t=1}^T$ and $\{ \mathbf h^b_t \}_{t=1}^T$ are the forward and backward hidden states of the BLSTM, respectively:
\begin{align}
\label{eq:blstm}
    \mathbf h^f_t &= \mbox{LSTM} (\mathbf x_t , \mathbf h^f_{t-1}  ), \\
    \mathbf h^b_t &= \mbox{LSTM} (\mathbf x_t , \mathbf h^b_{t+1}  ).
\end{align}%
We also use the layer normalization \cite{Jimmy-arxiv-2016}.

\textbf{Semantic Attention.}
The input and output semantic attention of this model is almost identical to those of the captioning model in section \ref{subsec:model_description},
only except that the word representation $\mathbf c_t \in \mathbb R^d$ is used as input  at each time step, instead of previous word vector $\mathbf y_{t-1}$. %
Then the attention weighted word vector $\{ \mathbf x_t \}_{t=1}^T$ is fed into the BLSTM. %

The output semantic attention is also similar to that of the captioning model in section \ref{subsec:model_description},
only except that we apply the attention only once at $t$-th step where the \texttt{<blank>} token is taken as input.
We feed the output of the BLSTM
\begin{align}
    \mathbf o_t = \mathrm{tanh}( \mathbf W_o [ \mathbf h_{t}^f ; \mathbf h_{t}^b ] + \mathbf b_o ),
\end{align}
where $\mathbf W_o \in \mathbb R^{D \times 2D}$ and $\mathbf b_o \in \mathbb R^D$,
into the output attention function $\mathbf \varphi$,
which generates $\mathbf p \in \mathbb R^D$ as in Eq.(\ref{eq:attention_out_vec}) of the description model,
$\mathbf p = \varphi(\mathbf o_t, \{\mathbf a_i\})$.

Finally, the output word probability $\mathbf y$ given $\{ \mathbf c_t \}_{t=1}^T$ is obtained via
softmax on $\mathbf p$ as
\begin{align}
    p(\mathbf y \mid \{ \mathbf c_t \}_{t=1}^T)      &= \mathrm{softmax} ( \mathbf W_y \mathbf p + \mathbf b_y),
\end{align}%
where parameters include $\mathbf W_y \in \mathbb R^{|\mathcal V| \times D}$ and $\mathbf b_y \in \mathbb R^{|\mathcal V|}$.

\textbf{Training.}
During training, we minimize the loss $\mathcal L$ as
\begin{align}
\label{eq:loss_function_2}
\mathcal{L} = - \log p(\mathbf y) + \lambda_1 (g(\mathbf \beta) + g(\mathbf \gamma)) + \lambda_2 \mathcal{L}_{con},
\end{align}%
where $\lambda_1, \lambda_2$ are hyperparameters, and $g$ is the same regularization function of Eq.(\ref{eq:regularizer}).
Again, $\mathcal{L}_{con}$ is the cost of the concept word detector in Eq.(\ref{eq:concept_cost}).

\section{Experiments}
\label{sec:experiments}

\begin{table*}[tb]
\setlength\tabcolsep{5pt} %
\centering
\small
\newcommand{\ranked}[1]{\xspace\scriptsize\sf{(#1)}}
\begin{tabular}{|c|ccccccc|}
\hline
{\footnotesize Movie Description}
                                  & B1                       & B2                       & B3                       & B4                       & M                        & R                        & Cr                       \\ \hline
EITanque \cite{kaufman-arxiv-2016} & 0.144\ranked{4}          & 0.042\ranked{5}          & 0.016\ranked{3}          & 0.007\ranked{2}          & 0.056\ranked{7}          & 0.130\ranked{7}          & 0.098\ranked{2}          \\
S2VT \cite{venugopalan-iccv-2015} & \textbf{0.162}\ranked{1} & \textbf{0.051}\ranked{1} & \textbf{0.017}\ranked{1} & 0.007\ranked{2}          & 0.070\ranked{4}          & 0.149\ranked{4}          & 0.082\ranked{4}          \\
SNUVL                             & 0.157\ranked{2}          & 0.049\ranked{2}          & 0.014\ranked{4}          & 0.004\ranked{6}          & 0.071\ranked{2}          & 0.147\ranked{5}          & 0.070\ranked{6}          \\
sophieag                          & 0.151\ranked{3}          & 0.047\ranked{3}          & 0.013\ranked{5}          & 0.005\ranked{4}          & \textbf{0.075}\ranked{1} & 0.152\ranked{2}          & 0.072\ranked{5}          \\
ayush11011995                     & 0.116\ranked{8}          & 0.032\ranked{7}          & 0.011\ranked{7}          & 0.004\ranked{6}          & 0.070\ranked{4}          & 0.138\ranked{6}          & 0.042\ranked{8}          \\
rakshithShetty                    & 0.119\ranked{7}          & 0.024\ranked{8}          & 0.007\ranked{8}          & 0.003\ranked{8}          & 0.046\ranked{8}          & 0.108\ranked{8}          & 0.044\ranked{7}          \\
Aalto                             & 0.070\ranked{9}          & 0.017\ranked{9}          & 0.005\ranked{9}          & 0.002\ranked{9}          & 0.033\ranked{9}          & 0.069\ranked{9}          & 0.037\ranked{9}          \\ \hline
Base-SAN                          & 0.123\ranked{6}          & 0.038\ranked{6}          & 0.013\ranked{5}          & 0.005\ranked{4}          & 0.066\ranked{6}          & 0.150\ranked{3}          & 0.090\ranked{3}          \\
CT-SAN                            & 0.135\ranked{5}          & 0.044\ranked{4}          & \textbf{0.017}\ranked{1} & \textbf{0.008}\ranked{1} & 0.071\ranked{2}          & \textbf{0.159}\ranked{1} & \textbf{0.100}\ranked{1} \\ \hline
\end{tabular}
\hfill
\begin{tabular}{|l|c|}
\hline
Fill-in-the-Blank  & {\footnotesize Accuracy} \\ \hline
Simple-LSTM        & 30.9                     \\
Simple-BLSTM       & 31.6                     \\
Base-SAN (Single)  & 34.5                     \\ \hline
Merging-LSTM \cite{mazaheri-arxiv-2016}   & 34.2                     \\
Base-SAN (Ensemble)& 36.9                     \\ \hline
SNUVL  (Single)    & 38.0                     \\
SNUVL  (Ensemble)  & 40.7                     \\ \hline
CT-SAN (Single)    & 41.9                     \\
CT-SAN (Ensemble)  & \textbf{42.7}            \\
\hline
\end{tabular}
\medskip
\caption{
    \textbf{Left}:
    Performance comparison for the movie description task on the LSMDC 2016 public test dataset.
    For language metrics, we use BLEU (B), METEOR (M), ROUGE (R), and CIDEr (Cr). We also show the ranking in parentheses.
    \textbf{Right}:
    Accuracy comparison (in percentage) for the movie fill-in-the-blank task.
}
\label{tbl:results_vidcap}
\vspace{-7pt}
\end{table*}

We report the experimental results of the  proposed models for the four tasks of LSMDC 2016.
More experimental results and implementation details can be found in the supplementary file.

\subsection{The LSMDC Dataset and Tasks}
\label{sec:lsmdc_dataset}

The LSMDC 2016 comprises four video-to-language tasks on the LSMDC dataset,
which contains a parallel corpus of 118,114 sentences and 118,081 video clips sampled from 202 movies. %
We strictly follow the evaluation protocols of the challenge.
We defer more details of the dataset and challenge rules to \cite{rohrbach-ijcv-2017} and the challenge homepage\footnote{\url{https://sites.google.com/site/describingmovies/}.}.

\smallskip
\textbf{Movie Description}.
This task is related to video captioning; given a short video clip, its goal is to generate a single descriptive sentence. %
The challenge provides a subset of LSMDC dataset named \textit{LSMDC16}. %
It is divided into training, validation, public test, and blind test set, whose sizes are 91,941, 6,542, 10,053, and 9,578, respectively.
The official performance metrics include BLEU-1,2,3,4 \cite{Papineni-acl-2002}, METEOR \cite{Banerjee-acl-2005}, ROUGE-L \cite{Lin-was-2004} and CIDEr \cite{Vedantam-arxiv-2014}.

\textbf{Multiple-Choice Test}.
Given a video query and five candidate captions, from which its goal is to find the best option.
The correct answer is the GT caption of the query video, and four other distractors are randomly chosen from the other captions that have different activity-phrase labels from the correct answer.
The evaluation metric is the percentage of correctly answered test questions out of 10,053 public-test data.

\textbf{Movie Retrieval}.
The objective is, given a short query sentence, to search for its corresponding video out of 1,000 candidate videos, sampled from the LSMDC16 public-test data.
The evaluation metrics include Recall@1/5/10, and Median Rank (MedR).
The Recall@$k$ means the percentage of the GT video included in the first $k$ retrieved videos,
and the MedR indicates the median rank of the GT.
Each algorithm predicts $1,000\times 1,000$ pairwise rank scores between phrases and videos, from which all the evaluation metrics are calculated.

\textbf{Movie Fill-in-the-Blank}.
This task is related to visual question answering;
given a video clip and a sentence with a blank in it, its goal is to predict a single correct word to fill in the blank.
The test set includes 30,000 examples from 10,000 clips (\ie about 3 examples per sentence).
The evaluation metric is the prediction accuracy, which is the percentage of predicted words that match with GTs.

\subsection{Quantitative Results}
\label{sec:quant_results}

We compare with the results on the public dataset in the official evaluation server of LSMDC 2016 as of the submission deadline
(\ie November 15th, 2016 UTC 23:59).
Except award winners, the LSMDC participants have no obligation to disclose their identities or used technique. 
Below we use the IDs in the leaderboard to denote participants. 

\textbf{Movie description.}
Table \ref{tbl:results_vidcap} compares the performance of movie description between different algorithms.
Among comparable models, our approach ranks (5, 4, 1, 1)-th in the BLEU language metrics,
and (2, 1, 1)-th in the other language metrics.
That is, our approach ranks first in four metrics, which means that our approach is comparable to the state-of-the-art methods.
In order to quantify the improvement by the proposed concept word detection and semantic attention,
we implement a variant (Base-SAN), which is our model of Fig.\ref{fig:model_vidcap}  without those two components.
As shown in Table \ref{tbl:results_vidcap}, the performance gaps between (CT-SAN) and (Base-SAN) are significant.

\begin{table}[t!]
\centering
\small
\newcommand{\ranked}[1]{$^\text{(#1)}$}
\setlength\tabcolsep{1.5pt}
\begin{tabular}{|l|c|cccc|}
\hline
\multicolumn{1}{|c|}{Tasks }           & {\footnotesize Multiple-Choice} & \multicolumn{4}{c|}{\footnotesize Movie Retrieval} \\ \hline
\multicolumn{1}{|c|}{Methods}          & {Accuracy}                      & {\footnotesize R@1}                                 & {\footnotesize R@5 } & {\footnotesize R@10 } & {\footnotesize MedR } \\ \hline
Aalto                                  & 39.7          & --           & --            & --            & --          \\
SA-G+SA-FC7 \cite{torabi-arxiv-2016}   & 55.1          & 3.0          & 8.8           & 13.2          & 114         \\
LSTM+SA-FC7 \cite{torabi-arxiv-2016}   & 56.3          & 3.3          & 10.2          & 15.6          & 88          \\
C+LSTM+SA-FC7 \cite{torabi-arxiv-2016} & 58.1          & 4.3          & 12.6          & 18.9          & 98          \\
Base-SAN (Single)                      & 60.1          & 4.3          & 13.0          & 18.2          & 83          \\ %
Base-SAN (Ensemble)                    & 64.0          & 4.4          & 13.9          & 19.3          & 74     \\ %
SNUVL (Single)                         & 63.1          & 3.8          & 13.6          & 18.9          & 80          \\ %
EITanque \cite{kaufman-arxiv-2016}     & 63.7          & 4.7          & 15.9          & 23.4          & 64          \\
SNUVL (Ensemble)                       & 65.7          & 3.6          & 14.7          & 23.9          & 50          \\ \hline
CT-SAN (Single)                        & 63.8          & 4.5          & 14.1          & 20.9          & 67          \\ %
CT-SAN (Ensemble)                      & \textbf{67.0} & \textbf{5.1} & \textbf{16.3} & \textbf{25.2} & \textbf{46} \\ %
\hline
\end{tabular}

\medskip
\caption{
    Performance comparison for the multiple-choice test (accuracy in percentage)
    and movie retrieval task: Recall@k (R@k, higher is better) and Median Rank (MedR, lower is better).
}
\label{tbl:results_mc_ret}
\vspace{-7pt}
\end{table}

\begin{figure*}[t!]
\centering
\includegraphics[width=0.97\textwidth]{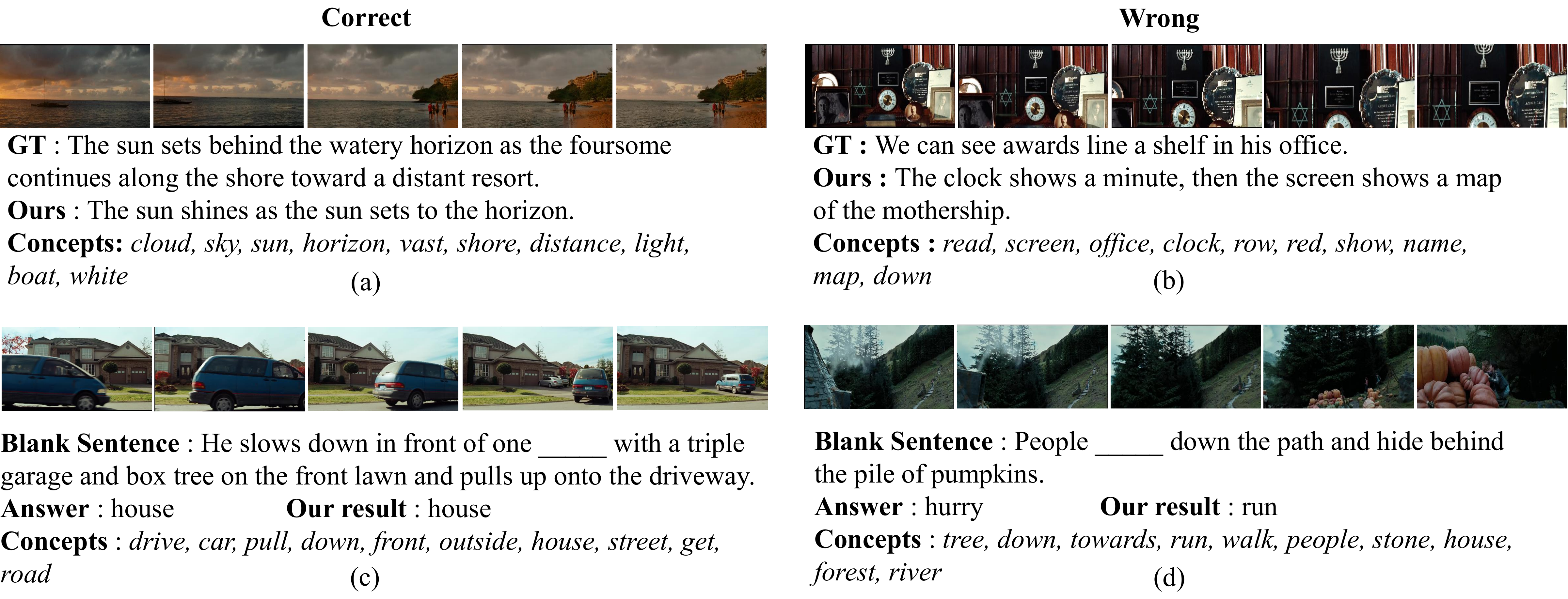}
\vspace{3pt}

\includegraphics[width=0.97\textwidth]{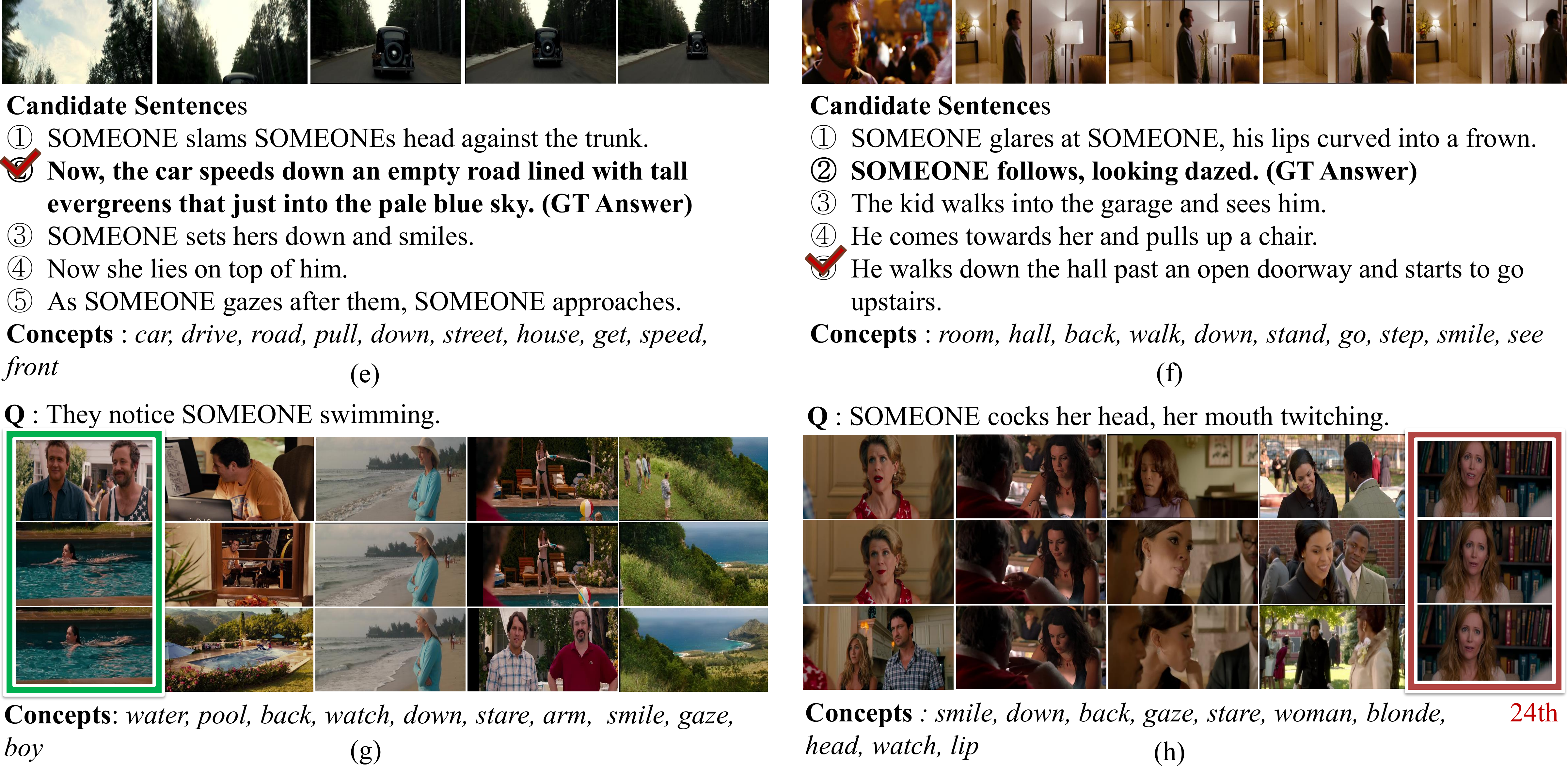}
\vspace{3pt}
\caption{Qualitative examples of the four vision-to-language tasks:
    (a)-(b) movie description, (c)-(d) fill-in-the-blank, (e)-(f) multiple-choice, and (g)-(h) movie retrieval.
    The left column shows correct examples while the right column shows wrong examples.
    In (h), we also show our retrieval ranks of the GT clips (the red box), 24th.
    We present more, clearer, and larger examples in the supplementary file.
}
\vspace{-3pt}
\label{fig:examples}
\end{figure*}

\textbf{Movie Fill-in-the-Blank.}
Table \ref{tbl:results_vidcap} also shows the results of the fill-in-the-blank task.
We test an ensemble of our models, denoted by (CT-SAN) (Ensemble);
the answer word is obtained by averaging the output word probabilities of three identical models trained independently. %
Our approach outperforms all the participants with large margins.
We also compare our model with a couple of baselines:
(CT-SAN) outperforms
the simple single-layer LSTM/BLSTM variants with the scoring layer on top of the blank location,
and (Base-SAN), which is the base model of (CT-SAN) without the concept detector and semantic attention.

\textbf{Movie Multiple-Choice Test.}
For the multiple-choice test, our approach also ranks first as shown in Table \ref{tbl:results_mc_ret}.
As in the fill-in-the-blank, the multiple-choice task also benefits from
the concept detector and semantic attention.
Moreover, an ensemble of six models trained independently further improves the accuracy from 63.8\% to 67.0\%. %

\textbf{Movie Retrieval.}
Table \ref{tbl:results_mc_ret} compares Recall@$k$ (R@k) and Median Rank (MedR) metrics between different methods. %
We also achieve the best retrieval performance with significant margins from baselines.
Our (CT-SAN) (Ensemble) obtains the video-sentence similarity matrix with an ensemble of two different models.
First, we train six retrieval models with different parameter initializations.
Second, we obtain the similarity matrix using the multiple-choice version of (CT-SAN), because it can also generate a similarity score for a video-sentence pair.
Finally, we average the seven similarity matrices into the final similarity matrix.

\smallskip
\subsection{Qualitative Results}
\label{sec:qual_results}

Fig.\ref{fig:examples} illustrates qualitative results of our algorithm with correct or wrong examples for each task.
In each set, we show sampled frames of a query video, groundtruth (GT), our prediction (Ours), and the detected concept words.
We provide more examples in the supplementary file.

\textbf{Movie Description.}
Fig.\ref{fig:examples}(a)-(b) illustrates examples of our movie description.
The predicted sentences are often related to the content of clips closely,
but the words themselves are not always identical to the GTs.
For instance, the generated sentence for Fig.\ref{fig:examples}(b) reads \textit{the clock shows a minute},
which is relevant to the video clip although its GT sentence much focuses on \textit{awards on a shelf}.
Nonetheless, the concept words relevant to the GT sentence are well detected such as \textit{office} or \textit{clock}.

\textbf{Movie Fill-in-the-Blank.}
Fig.\ref{fig:examples}(c) shows that the detected concept words are well matched with the content of the clip, and possibly help predict  the correct answer.
Fig.\ref{fig:examples}(d) is a near-miss case where our model also predict a plausible answer (\eg \textit{run} instead of \textit{hurry}).

\textbf{Movie Multiple-Choice Test.}
Fig.\ref{fig:examples}(e) shows that our concept detection successfully guides the model to select the correct answer.
Fig.\ref{fig:examples}(f) is an example of failure to understand the situation;
the fifth candidate is chosen because it is overlapped with much of detected words such as \textit{hall}, \textit{walk}, \textit{go}, although the correct answer is the second.

\textbf{Movie Retrieval.}
Interestingly, the concept words of Fig.\ref{fig:examples}(g) capture the abstract relation between \textit{swimming}, \textit{water}, and \textit{pool}.
Thus, the first to fifth retrieved clips include \textit{water}.
Fig.\ref{fig:examples}(h) is a near-miss example in which our method fails to catch rare word like \textit{twitch} and \textit{cocks}.
The first to fourth retrieved clips contain a woman's head and mouth, yet miss to catch subtle movement of mouth.

\section{Conclusion}
\label{sec:conclusion}
\vspace{-2pt}
We proposed an end-to-end trainable approach for detecting a list of concept words that can be used as semantic priors for multiple-video-to-language models.
We also developed a semantic attention mechanism that effectively exploits the discovered concept words.
We implemented our approach into multiple video-to-language models to participate in four tasks of LSMDC 2016.
We demonstrated that our method indeed improved the performance of video captioning, retrieval, and question answering, and finally
won three tasks in LSMDC 2016, including\, \textit{fill-in-the-blank}, \textit{multiple-choice test}, and \textit{movie retrieval}.

\medskip\noindent\textbf{Acknowledgements}.
This research is partially supported by Convergence Research Center through National Research Foundation of Korea (2015R1A5A7037676).
Gunhee Kim is the corresponding author.

{\small
\bibliographystyle{ieee}
\bibliography{cvpr17_lsmdc}
}

\clearpage
\appendix

\maketitle\thispagestyle{empty}

{\noindent\bf\Large Appendix}
\bigskip

\section{Details of Video-to-Language Models}
\label{sec:supp_v2l_models}

\begin{figure*}%
\centering
\includegraphics[trim=0cm 0.05cm 0cm 0cm,clip,width=0.99\textwidth]{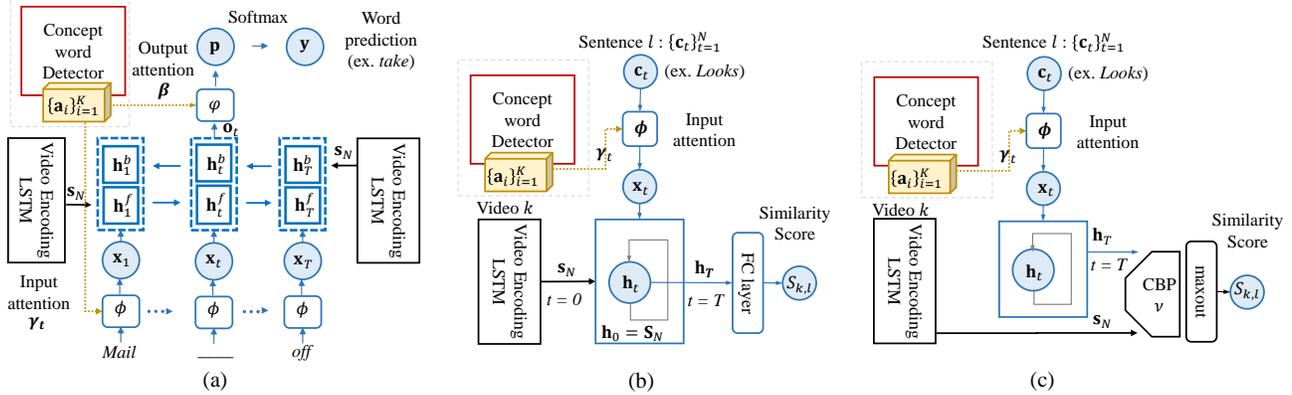}
\vspace{5pt}
\caption{
    (Repeat of Figure \ref{fig:models_fib_mc_ret} in the main paper)
    The model architectures for
    (a) fill-in-the-blank (section \ref{subsec:model_blank}),
    (b) multi-choice (section \ref{subsec:supp_model_choice}),
    and (c) movie retrieval task (section \ref{subsec:supp_model_retrieval}).
Each of the models take advantage of the concept word detector described
illustrated in Figure \ref{fig:model_vidcap},
and semantic attention for the sake of its objective.
}

\label{fig:supp_models_fib_mc_ret}
\vspace{-5pt}
\end{figure*}

In this section, we describe the further details of video-to-language models (section \ref{sec:v2l_models}).

\smallskip
\subsection{A Model for Multiple-Choice Test}
\label{subsec:supp_model_choice}

Figure \ref{fig:supp_models_fib_mc_ret}(b) illustrates the proposed model for the multiple-choice test.
It takes a video and five choice sentences among which only one is the correct answer.
Hence, our model computes the compatibility scores between the query video and five sentences,
and selects the one with the highest score.

The multiple-choice model shares much resemblance to the model for fill-in-the-blank in Figure \ref{fig:supp_models_fib_mc_ret}(a).
First, it is based on the LSTM network, although it is not bi-directional.
Second, it inputs the query video into the video encoding LSTM, and use its last hidden state $\mathbf s_N$ to initialize the following LSTM.
Third, it uses the same word representation $\{ \mathbf c_t \}_{t=1}^T$ for each candidate sentence.
Finally, it exploits the same \textit{input semantic attention} of Eq.(\ref{eq:attention_input_wgt})--(\ref{eq:attention_input_vector}),
although it does not apply the \textit{output semantic attention} because output is not a word but a score in this task.

We obtain a joint embedding of a pair of a single video and a sentence using the LSTM network:
\begin{align}
    \mathbf h_t = \mathrm{LSTM}(\mathbf x_t, \mathbf h_{t-1})
\end{align}
where $\x_t = \phi (\mathbf c_t, \{\mathbf a_i \}) \in \mathbb R^D$
is obtained via the input semantic attention $\phi$ of Eq.(\ref{eq:attention_input_wgt})--(\ref{eq:attention_input_vector}),
from the input sentence representation $\{\mathbf c_t\}_{t=1}^{T}$.
We also initialize the hidden state $\mathbf h_0 = \mathbf s_N$ by the final hidden state of video representation.
Once the sentence is fed into the LSTM, we obtain a multimodal embedding of a video-sentence pair as the final hidden state $\mathbf h_T$ of the LSTM.

\textbf{Alignment Objective.}
The objective of the multiple-choice model is to assign high scores for the correctly matched video-sentence pairs but low scores for incorrect pairs.
Therefore, we predict a similarity score $S_{kl}$ between a movie clip $k$ and a sentence $l$ as follows:
\begin{align}
    \label{eq:mc_score_layer}
    S_{kl} = \mathbf (\mathbf W_s)^\top \mbox{ReLU}( \mathbf W_a \mathbf h_T + \mathbf b_a),
\end{align}
where $\mathbf W_a \in \mathbb R^{D \times D}$, $\mathbf b_a \in \mathbb R^D$ and $\mathbf W_s \in \mathbb R^{D}$
are parameters.
We train the model using a max-margin structured loss objective:
\begin{align}
    \mathcal{L} = & \sum_k \sum_{l=1}^{5} \max(0, S_{k,l} - S_{k,l^*} + \Delta) \nonumber
        \\ & +\lambda_1 \cdot g(\mathbf \gamma) + \lambda_2 \mathcal{L}_{con}
    \label{eq:mc_maxmarginloss}
\end{align}
where $l^*$ denotes the answer sentence among the five candidates.
This objective encourages a positive video-sentence pair to have a higher score than a misaligned negative pair by a margin $\Delta$.
We use $\Delta = 1$ in our experiments.

At test, for a query video $k$, we compute five scores $\{S_{k,l}\}_{l=1}^5$ of the candidate sentences,
and select the one with maximum score $S_{k,l}$ as the answer.

\smallskip
\subsection{A Model for Retrieval}
\label{subsec:supp_model_retrieval}

Figure \ref{fig:supp_models_fib_mc_ret}(c) illustrates our model for movie retrieval.
The basic idea is to compute a score for a query text and video pair,
by learning a joint representation between two modalities (\ie query text and video)
using the CBP (Compact Bilinear Pooling) layer \cite{akira-emnlp-2016}.

For the video encoding, we use the final hidden state $\mathbf s_N$ of the video encoding LSTM as done in other models.
We also obtain a query representation via input semantic attention like as in section \ref{subsec:supp_model_choice}, through the LSTM network:
\begin{align}
    \mathbf h_t = \mathrm{LSTM}(\mathbf x_t, \mathbf h_{t-1})
\end{align}
Similarly,
$\x_t = \phi (\mathbf c_t, \{\mathbf a_i \}) \in \mathbb R^D$
is obtained via the input semantic attention of Eq.(\ref{eq:attention_input_wgt})--(\ref{eq:attention_input_vector}),
from the input query sentence representation $\{\mathbf c_t\}_{t=1}^{T}$.
Then, we use the final hidden state $\mathbf h_T$ of query encoding LSTM as query representation. %

To measure a similarity score $S_{k,l}$ between a movie $k$ and a sentence $l$ as follows (see Figure \ref{fig:supp_models_fib_mc_ret}(c)):
\begin{align}
    \label{eq:qv_score}
    S_{k,l} = \mathbf (\mathbf W_s)^\top \mbox{maxout}( \mathbf W_p^\top \nu(\mathbf s_N, \mathbf h_T))
\end{align}
where $\nu(\cdot)$ denotes the CBP (Compact Bilinear Pooling) layer \cite{akira-emnlp-2016}, which captures the interactions between different modalities better than simple concatenation.
That is, we learn the multimodal space for common features between video encoding LSTM and query encoding LSTM.
The joint representation extracted from the MCB layer is multiplied by $\mathbf W_p \in \mathbb R^{8,000 \times 1,500}$, %
and further processed by a consequent maxout layer~\cite{Goodfellow-icml-2013},
which yields non-sparse activations while mitigating overfitting.
Finally, we obtain the score $S_{k,l}$ by multiplying the output by $\mathbf W_s \in \mathbb R^{1500 \times 1}$.

We use the same max-margin structured loss objective with the multiple-choice model:
\begin{align}
    \label{eq:maxmarginloss_ret}
    \mathcal{L} =& \sum_k \sum_{l} \max(0, S_{k,l} - S_{k,l^*} + \Delta)     \nonumber \\
    & ~ + \lambda_1 \cdot g(\gamma) + \lambda_2 \mathcal{L}_{con}
\end{align}
which encourages a positive video-sentence pair to have a higher score than a misaligned pair by a margin $\Delta$ (\eg $\Delta = 3$ in our experiments).

At test, for a query sentence $k$, we compute scores $\{S_{k,l}\}_{l}$ for all videos $l$ in the test set.
From the score matrix, we can rank the videos for the query.
As mentioned in section \ref{sec:quant_results}, an ensemble of multiple score matrices
is used in our final model, which yields much better retreival performance.

\begin{table*}[h!]
\centering
\small
\newcommand{\ul}[1]{{#1}}
\newcommand{\ranked}[1]{\xspace\scriptsize\sf{(#1)}}

\setlength\tabcolsep{7.0pt}
\begin{tabular}{|c|ccccccc|c|}
\hline
{\footnotesize Movie Description}
           & B1             & B2             & B3             & B4             & M              & R              & Cr             & FITB (Accuracy)     \\ \hline
rand-SAN   & 0.101          & 0.022          & 0.008          & 0.002          & 0.049          & 0.127          & 0.058          & 17.0     \\
no-ATT-SAN & 0.130          & 0.039          & 0.015          & 0.006          & 0.064          & 0.152          & 0.092          & 37.4     \\
Decoupled CT-SAN & \textbf{0.144} & \textbf{0.047} & \textbf{0.017} & 0.007& 0.066 & 0.152 & 0.086 & 35.4 \\
NN-SAN     & 0.122          & 0.035          & 0.012          & 0.005          & 0.058          & 0.142          & 0.078          & 12.1     \\
\hline
Base-SAN   & 0.123          & 0.038          & 0.013          & 0.005          & 0.066          & 0.150          & 0.090          & 34.5     \\
CT-SAN     & 0.135          & 0.044          & \textbf{0.017} & \textbf{0.008} & \textbf{0.071} & \textbf{0.159} & \textbf{0.100} & \textbf{41.9}  \\ \hline
\end{tabular}

\medskip
\caption{
    Performance comparison of more baselines, for the movie description task
    and for the fill-in-the-blank.
}
\label{tbl:supp_experiments_mf}
\vspace{-7pt}
\end{table*}

\begin{table}[h!]
\centering
\small
\newcommand{\ul}[1]{{#1}}
\newcommand{\ranked}[1]{\xspace\scriptsize\sf{(#1)}}

\setlength\tabcolsep{3.0pt}
\begin{tabular}{|l|c|cccc|}
\hline
\multicolumn{1}{|c|}{Tasks }           & {\footnotesize Multi-Choice}   & \multicolumn{4}{c|}{\footnotesize Movie Retrieval}                       \\ \hline
\multicolumn{1}{|c|}{Methods}          & {Accuracy}    & R@1          & R@5           & R@10          & MedR        \\ \hline
\ul{rand-SAN}                          & \ul{58.7}     & 2.1          &  8.1          & 11.6          & 104         \\ %
\ul{no-ATT-SAN}                        & \ul{61.1}     & 4.0          & 13.1          & 18.3          & 75          \\ %
\hline
Base-SAN                               & 60.1          & 4.3          & 13.0          & 18.2          & 83          \\ %
CT-SAN (Single)                        & 63.8          & 4.5          & 14.1          & 20.9          & 67          \\ %
CT-SAN (Ensemble)                      & \textbf{67.0} & \textbf{5.1} & \textbf{16.3} & \textbf{25.2} & \textbf{46} \\ %
\hline
\end{tabular}

\medskip
\caption{
    Performance comparison of more baselines on multiple-choice, and movie retrieval task.
}
\label{tbl:supp_experiments_mr}
\vspace{-7pt}
\end{table}

\begin{table*}[h!]
\centering
\small
\setlength\tabcolsep{7.0pt}
\begin{tabular}{|c|ccccccc|c|c|}
\hline
\multirow{3}*{} & \multicolumn{7}{|c|}{\footnotesize Movie Description}
                & {\footnotesize Fill-in-the-Blank}
                & {\footnotesize Multi-Choice}
\\ \cline{2-10}
                & B1             & B2             & B3             & B4             & M              & R              & Cr             & {Accuracy}    & {Accuracy}    \\ \hline
CT-SAN ($K=5$)  & 0.133          & 0.043          & 0.015          & 0.007          & 0.066          & 0.156          & 0.100          & 41.5          & 63.0          \\
CT-SAN ($K=10$) & 0.135          & \textbf{0.044} & \textbf{0.017} & \textbf{0.008} & \textbf{0.071} & \textbf{0.159} & 0.100          & \textbf{41.9} & \textbf{63.8} \\
CT-SAN ($K=20$) & \textbf{0.136} & \textbf{0.044} & 0.016          & \textbf{0.008} & 0.068          & 0.156          & \textbf{0.106} & \textbf{41.9} & 63.3          \\
\hline
\end{tabular}

\medskip
\caption{
    Performance comparison of our model (CT-SAN) in three tasks,
    varying the number of detected concept words $K$.
}
\label{tbl:supp_experiments_varyingK}
\vspace{-7pt}
\end{table*}

\begin{table*}[h!]
\centering
\small
\setlength\tabcolsep{7.0pt}
\begin{tabular}{|c|ccccccc|c|}
\hline
\multirow{2}*{} & \multicolumn{7}{|c|}{\footnotesize Movie Description}                                                                & {\footnotesize Fill-in-the-Blank} \\ \cline{2-9}
                & B1             & B2             & B3             & B4             & M              & R              & Cr             & {Accuracy}    \\ \hline
only input      & 0.128 & 0.041 & 0.012 & 0.006 & 0.066 & 0.151 & 0.078 & 37.7     \\
only output     & 0.130 & 0.043 & 0.014 & 0.005 & 0.067 & 0.148 & 0.097 & 39.1     \\
input\&output   & \textbf{0.135} & \textbf{0.044} & \textbf{0.017} & \textbf{0.008} & \textbf{0.071} & \textbf{0.159} & \textbf{0.100}& \textbf{41.9}    \\
\hline
\end{tabular}

\medskip
\caption{
    Performance comparison for ablation study of our model (CT-SAN) in two tasks. We apply the semantic attention to (i) only input, and (ii) only output.
}
\label{tbl:supp_experiments_ablation}
\vspace{-7pt}
\end{table*}

\section{Experimental Details}
\label{sec:supp_exp_details}

\subsection{Implementation Details}
\label{subsec:supp_implementation_details}

\textbf{Optimization.}
We train all of our models using the Adam optimizer \cite{kingma-iclr-2015} to minimize the loss, with an initial learning rate in the range of $10^{-4}$ to $10^{-5}$.
We adopt the data augmentation of image mirroring.
We also use batch shuffling in every training epoch.
We use Xavier initialization \cite{glorot-AISTATS-2010} for initializing the weight variables.
For all models, the LSTM (BLSTM) networks are two-layered in depth,
and we apply layer normalization \cite{Jimmy-arxiv-2016} and dropout \cite{srivastava-jmlr-2014} with a rate of 0.2
to reduce overfitting.

During training of fill-in-the-blank, multiple-choice, and retrieval models,
we initialize the parameters in the concept word detector component
with a pre-trained model of the movie description task.
The new parameters (\eg $\mathbf W_s, \mathbf W_a$ and the LSTM parameters for multi-choice test) are
initialized randomly,
and then the whole model is trained end-to-end using the provided training set.

\medskip

\textbf{Movie Description.}
The split of LSMDC16 dataset is provided by the challenge organizers:
(training, validation, test, blind test set) $= (101079, 9578, 10053, 7409)$ video-sentence pairs respectively.
We train our model using the training set of this split,
and the Para-Phrase AD sentences additionally provided by the challenge organizers.

\textbf{Fill-in-the-blank.}
The LSMDC16 dataset for the fill-in-the-blank is splitted into (training, validation, test set) $=(296961, 98483, 30350)$.
We also train our model using the officially provided training set only.
To improve prediction accuracy, we use an ensemble of models; the answer word is obtained
by averaging the output word probabilities of three copies of models trained with different initializations.

\textbf{Multiple-choice test.}
The training/validation/test split of LSMDC16 dataset is same as in the movie description task.
Although it is possible to include more negative sentences other than the provided four distractors
(we also find that it leads to a better accuracy), we experiment the models trained
using the four distractors only. %
we simply average the score matrix $S_{k, l}$ of individual models, to obtain the ensembled score matrix.
In our experiments, an ensemble of six copies of model trained independently,
denoted by CT-SAN (Ensemble), shows a considerable improvement of accuracy.

\textbf{Movie Retrieval.}
Our video encoding LSTM and query encoding LSTM use the same parameter setting with the LSTM networks for movie description.
We use the dropout~\cite{srivastava-jmlr-2014} before the maxout layer with the rate of 0.5.
The video-sentence similarity matrix $M \in \mathbb R^{1,000 \times 1,000}$ is obtained with an ensemble of identical models and multiple-choice model.
First, we train six retrieval models and one multiple-choice model with different parameter setting.
Second, we obtain the similarity matrix of alignment score from all possible pair between 1,000 natural language sentences and 1,000 movie clip.
To build an ensemble model, we average the multiple similarity matrices into the final similarity matrix.

\section{More Experimental Results}
\label{sec:supp_more_exps}

In this section, we provide additional experimental results to support the validity of
the proposed concept word detector and semantic attention models.

\subsection{On the Quality of Concept Words}

To study the effect of quality of concept words,
we present and experiment more baselines: (rand-SAN), (no-ATT-SAN), and (NN-SAN).

\smallskip
\textbf{Random Concept Words.}
A baseline (rand-SAN) is a variant of the same structure as (CT-SAN),
except that it uses \emph{random} concept words
instead of the ones detected by the concept word detector.
We uniformly sample $K = 10$ words for concept words,
from the $V$ candidates.

\smallskip
\textbf{Without Attention.}
We also study an effect of spatial attention in the proposed concept word detector (section \ref{sec:approach}).
With a simple baseline model denoted by {(no-ATT-SAN)},
the spatial attention component in the concept word detector
is replaced by a single two-layered LSTM.
Specifically, we compute the LSTM states $\{\mathbf h_n\}_{n=1}^N$
(a single LSTM instead of $L$ ones)
by feeding the average-pooled visual features $\overline{\mathbf v_n} \in \mathbb R^D$,
and then the concept confidence vector $\mathbf p$ using the last hidden state:
\begin{align}
    \mathbf h_n &= \mbox{LSTM}(\overline{\mathbf v_n}, \mathbf h_{n-1})           ~~(n = 1\ldots N), \\
    \mathbf p   &= \sigma (\mathbf W_p \mathbf h_N + \mathbf b_p)     ~~ \in \mathbb R^V,
\end{align}
which replaces Eq.(\ref{eq:concept_lstm2}) and Eq.(\ref{eq:concept_confidence_p}), respectively.
This baseline model simply transforms the video representation
into concept words, %
but does not involve any spatial attention.

\smallskip
\textbf{Nearest Neighbor.}
We also study a simpler baseline which use a nearest-neighbor method
instead of concept word detector.
This simple baseline is denoted by {(NN-SAN)}.
In this method, we simply take the concept words of the closest training video,
in terms of ResNet video features averaged over time.

\smallskip
\textbf{Quantitative Result.}{}
As shown in Table \ref{tbl:supp_experiments_mf} and \ref{tbl:supp_experiments_mr},
the performance of (no-ATT-SAN) is better than (Base-SAN) and (NN-SAN),
but poorer than the full model (CT-SAN),
in all of the four tasks.
This implies that the spatial attention helps detect concept words that are useful for video captioning.
Especially, (CT-SAN) outperforms (no-ATT-SAN) in the fill-in-the-blank and the multi-choice tasks with a large margin.
Nevertheless, using semantic attention turns out to be
more helpful than not using it,
as one can observe that (no-ATT-SAN) shows a better performance than (Base-SAN).

The performance of (rand-SAN) with semantic attention but with poor concept words,
is much inferior to (Base-SAN), which even lacks semantic attention.
As such, we find that the quality of concept words is crucial for performance enhancement.
Besides, retrieved words from (NN-SAN) are not so helpful in training semantic attention network.
(Decoupled CT-SAN) also shows worse performance than (CT-SAN) which is trained with an end-to-end manner.
These suggest that joint learning the concept word detector and the task-specific network
is effective in achieving a better performance.

\subsection{Ablation Study}

We conduct an additional ablation experiment on the semantic attention,
and present the results of movie description and FITB (Fill-in-the-Blank) tasks in Table \ref{tbl:supp_experiments_ablation}.

\subsection{On the Number of Concept Words}

We also conduct another simple experiment on the number of concept words.
We compute the performance of (CT-SAN), with changing the number of detected concept words, $K \in \{5, 10, 20\}$.
As shown in Table \ref{tbl:supp_experiments_varyingK},
we observe only a marginal performance difference.
However, as the number of concept words increases,
the time required to train the whole model increases,
and an overfitting is more prone to occur.

\begin{figure*}[p]
\centering
\includegraphics[width=0.75\textwidth,trim=0cm 0cm 0cm 0cm]{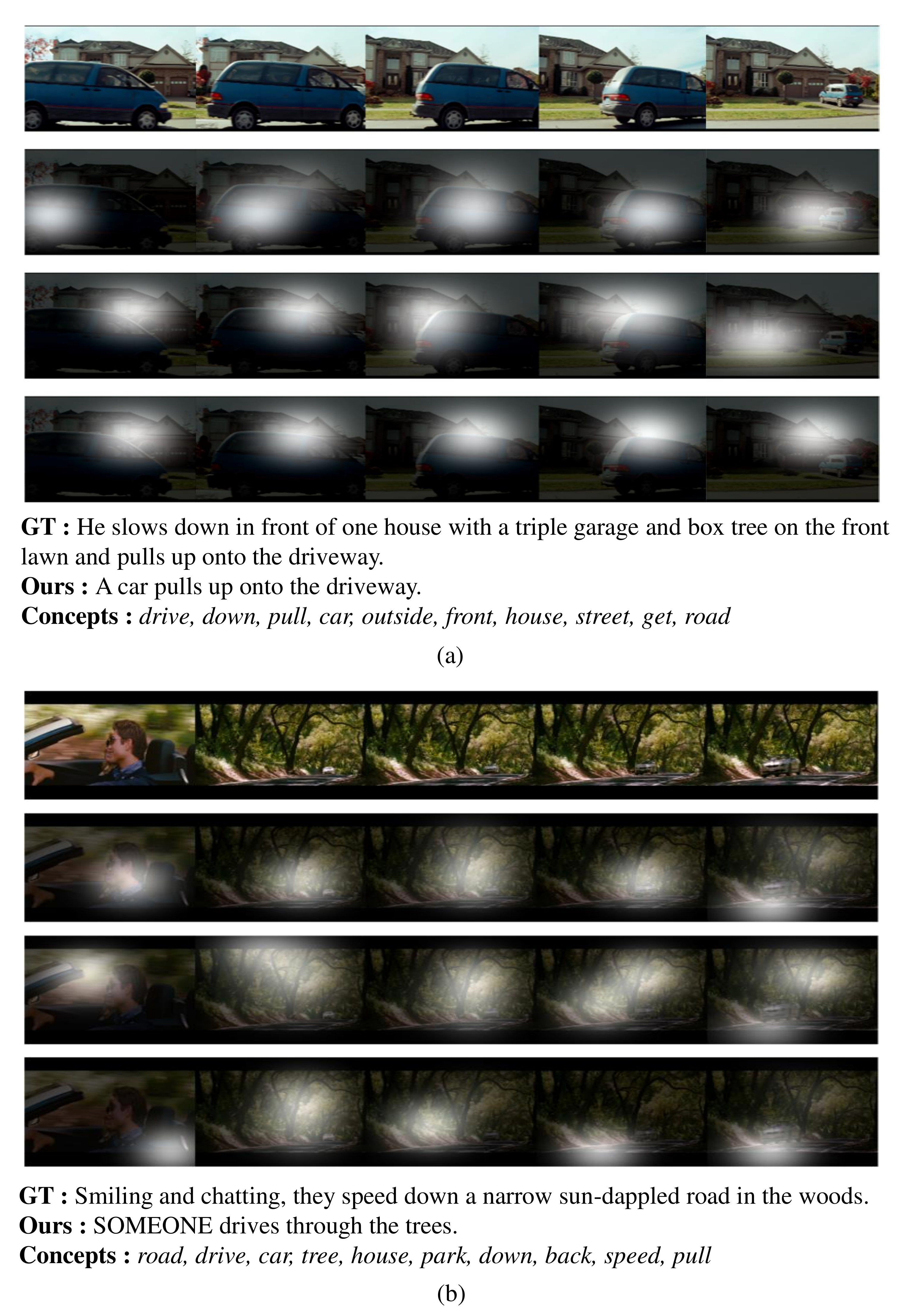}
\vspace{10pt}
\caption{Visualization of spatial attentions in the movie description model.
    In the first row, we show five sampled keyframes from the input movie.
    Below, we select three tracing-LSTMs among $L=16$ ones and show their spatial attention maps $\bm\alpha_t^{(l)}$ (see section \ref{sec:approach}).
}
\label{fig:example_attention1}
\vspace{-5pt}
\end{figure*}

\begin{figure*}[p]
\centering
\includegraphics[width=0.75\textwidth,trim=0cm 1cm 0cm 0cm]{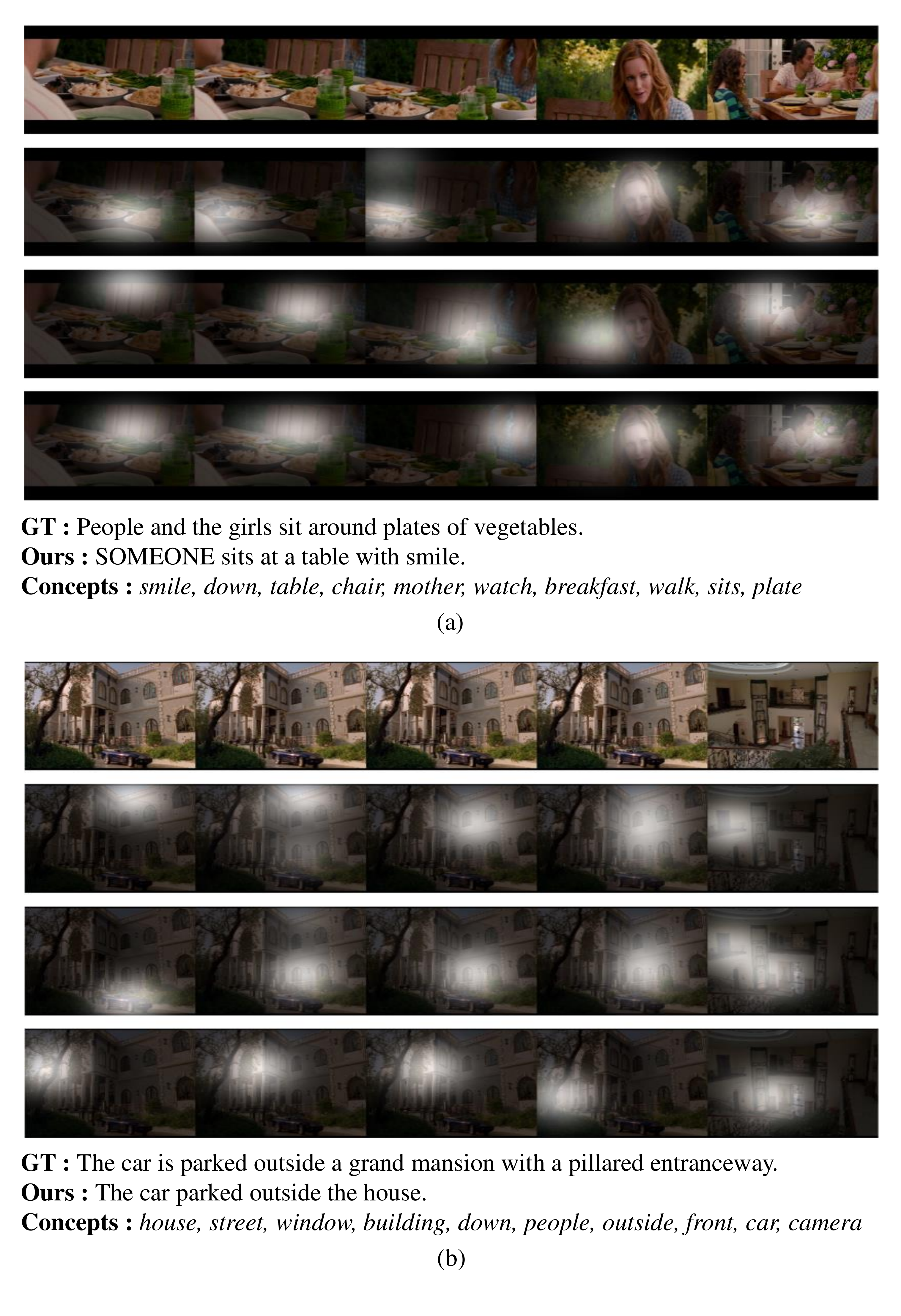}
\vspace{10pt}
\caption{Visualization of spatial attentions in the movie description model.
}
\label{fig:example_attention2}
\vspace{-5pt}
\end{figure*}

\begin{figure*}[p]
\centering
\includegraphics[width=0.99\textwidth,trim=0cm 0cm 0cm 0cm]{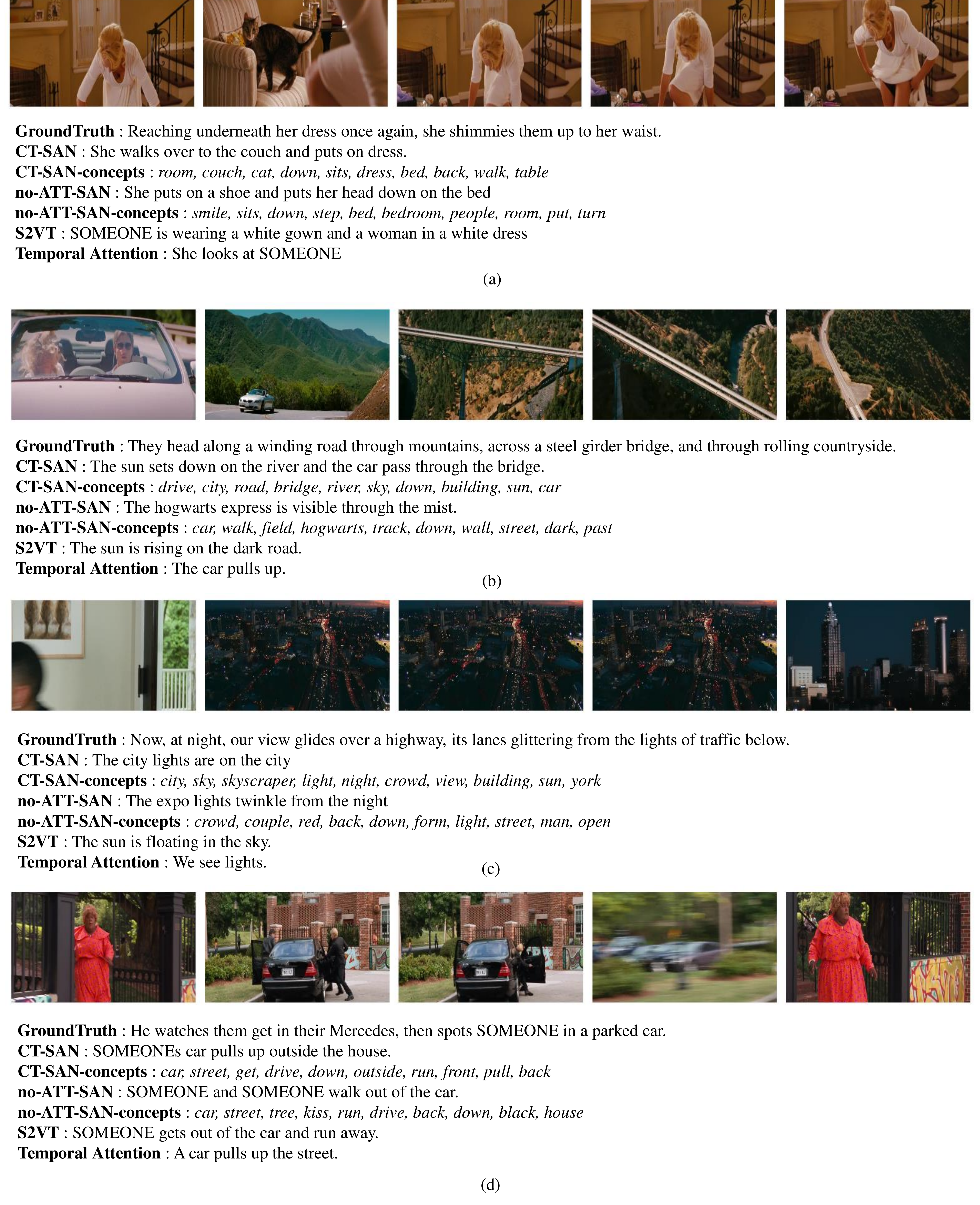}
\caption{Examples of our method and baselines in movie description.
    We show the generated description and the detected concept words of (CT-SAN) and (no-ATT-SAN).
    We also compare other movie description baselines, including
    S2VT \cite{venugopalan-iccv-2015} and Temporal Attention \cite{yao-iccv-2015} (we referenced their public code).
}
\label{fig:example_baseline1}
\vspace{-5pt}
\end{figure*}

\begin{figure*}[p]
\centering
\includegraphics[width=0.99\textwidth,trim=0cm 0cm 0cm 0cm]{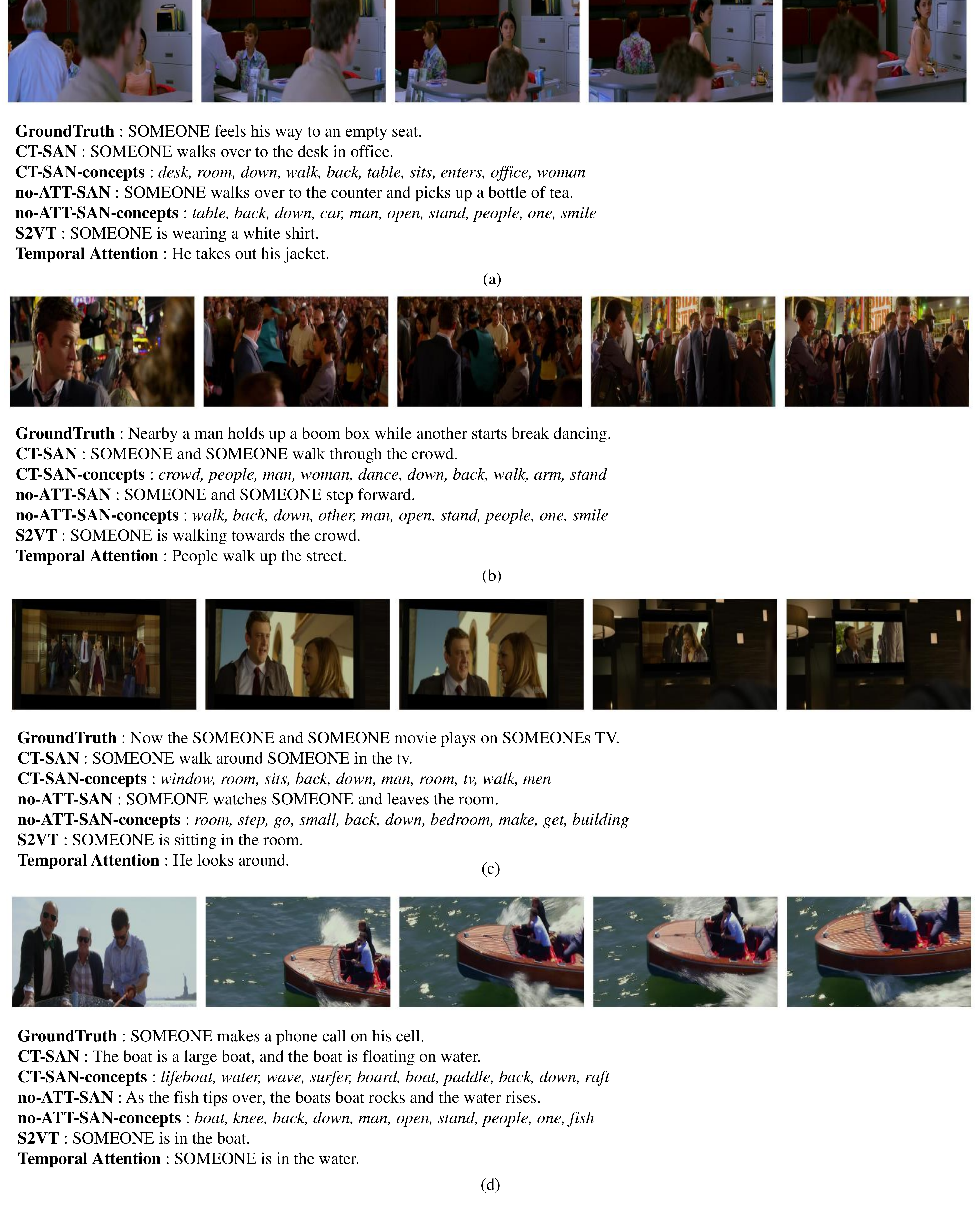}
\caption{More examples of our method and baselines in movie description.
}
\label{fig:example_baseline2}
\vspace{-5pt}
\end{figure*}

\section{More Examples and Qualitative Results}
\label{sec:supp_examples}

We visualize some examples of
the spatial attention computed in the concept word detector
in Figure \ref{fig:example_attention1}--\ref{fig:example_attention2}.
The spatial attentions roughly captures high-level concepts in the video
(\eg a blue \emph{car} moving left to right, in Fig.\ref{fig:example_attention1}(a)).
Figure \ref{fig:example_baseline1}--\ref{fig:example_baseline2}
show some examples of generated movie description with the concept words
detected by several baselines and our approach.

\begin{figure*}[p]
\centering
\includegraphics[width=0.95\textwidth,trim=0cm 0cm 0.8cm 0cm]{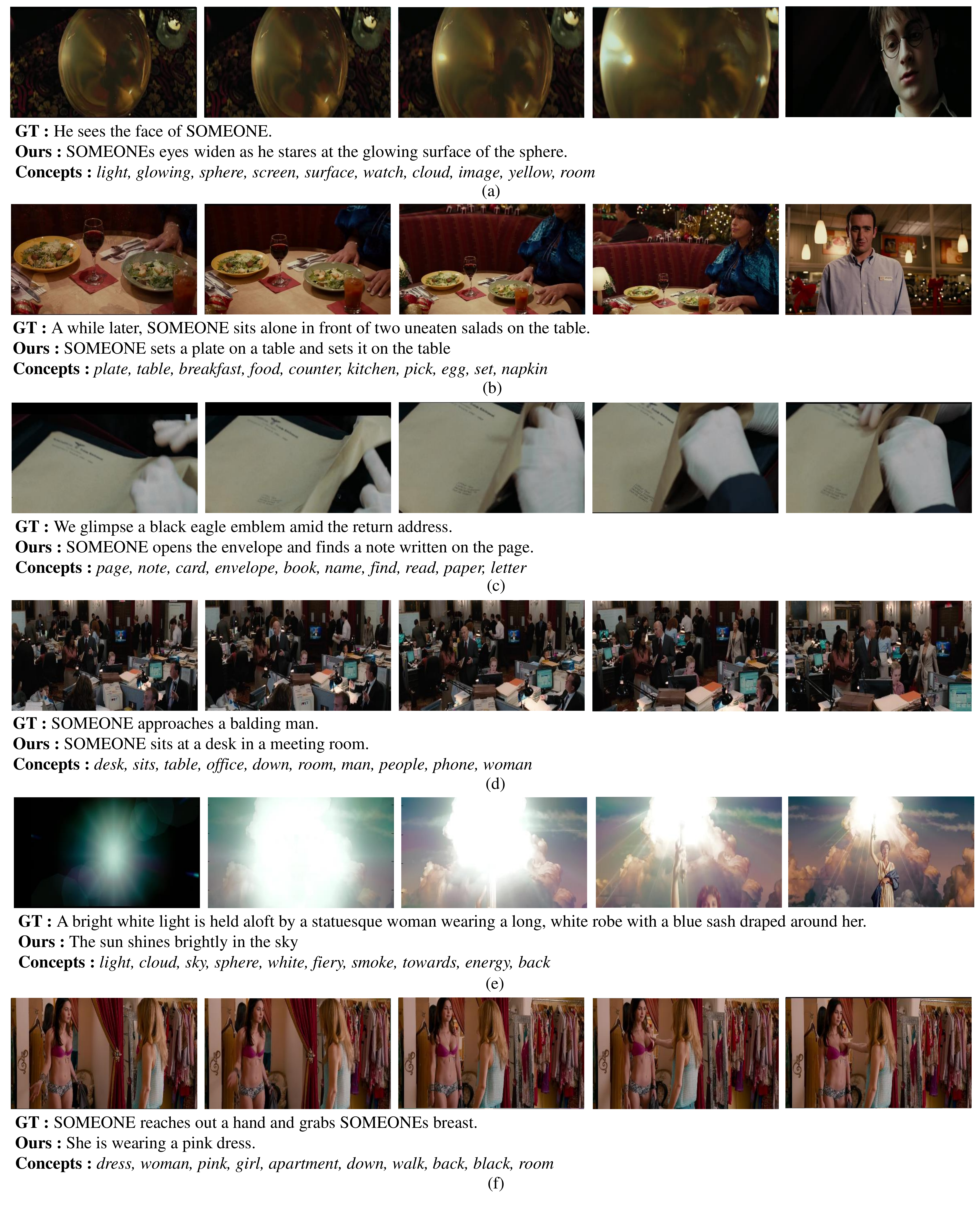}
\caption{Examples of movie descriptions. (a)-(d) are positive examples, and (e)-(f) are near-miss or wrong examples.
}
\label{fig:example_moviecap1}
\vspace{-5pt}
\end{figure*}

\begin{figure*}[p]
\centering
\includegraphics[width=0.95\textwidth,trim=0cm 0cm 0.8cm 0cm]{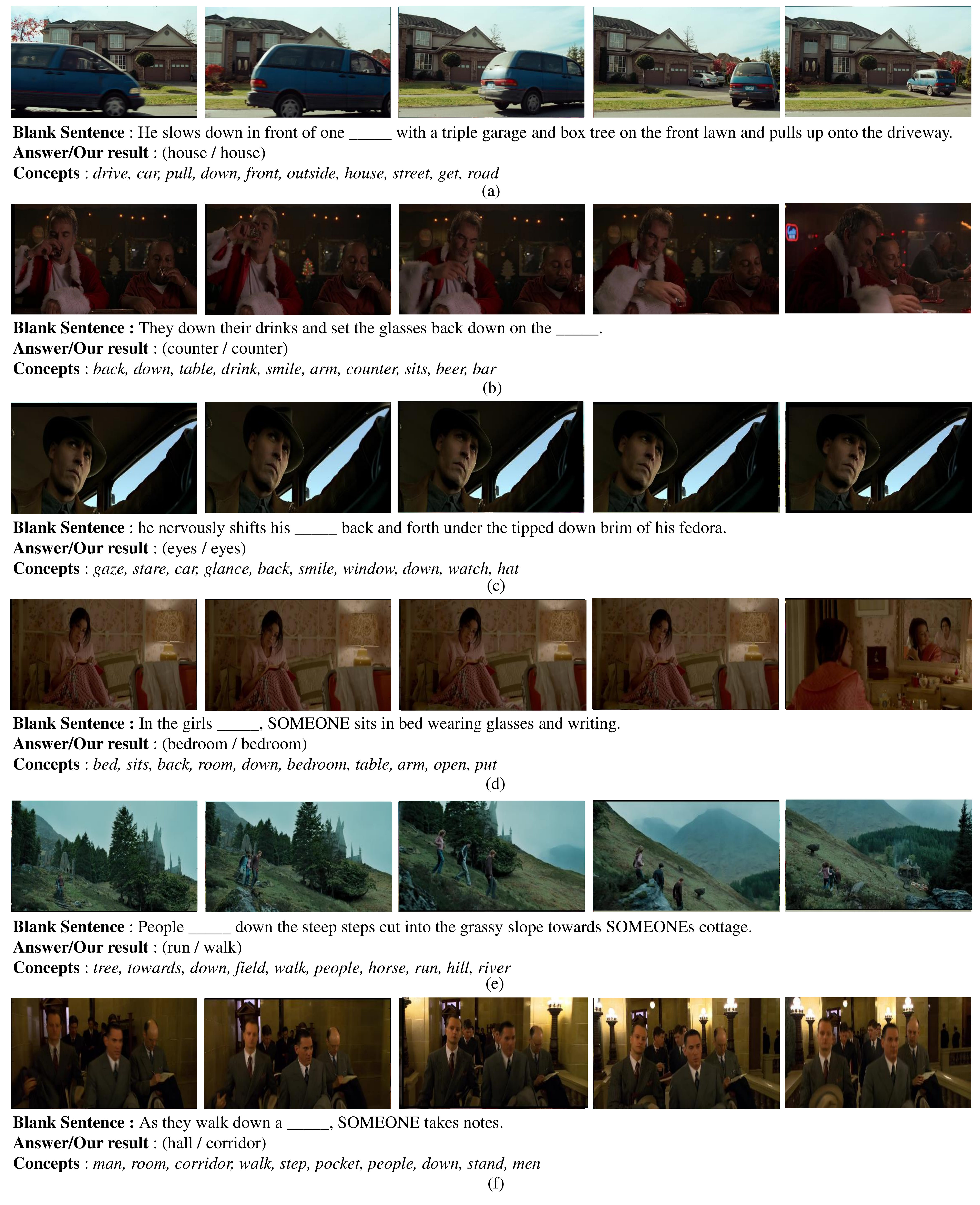}
\caption{Examples of fill-in-the-blank task. (a)-(d) are positive examples, and (e)-(f) are near-miss or wrong examples.
}
\label{fig:example_fib1}
\vspace{-5pt}
\end{figure*}

\begin{figure*}[p]
\centering
\includegraphics[width=0.95\textwidth,trim=0cm 0cm 0.8cm 0cm]{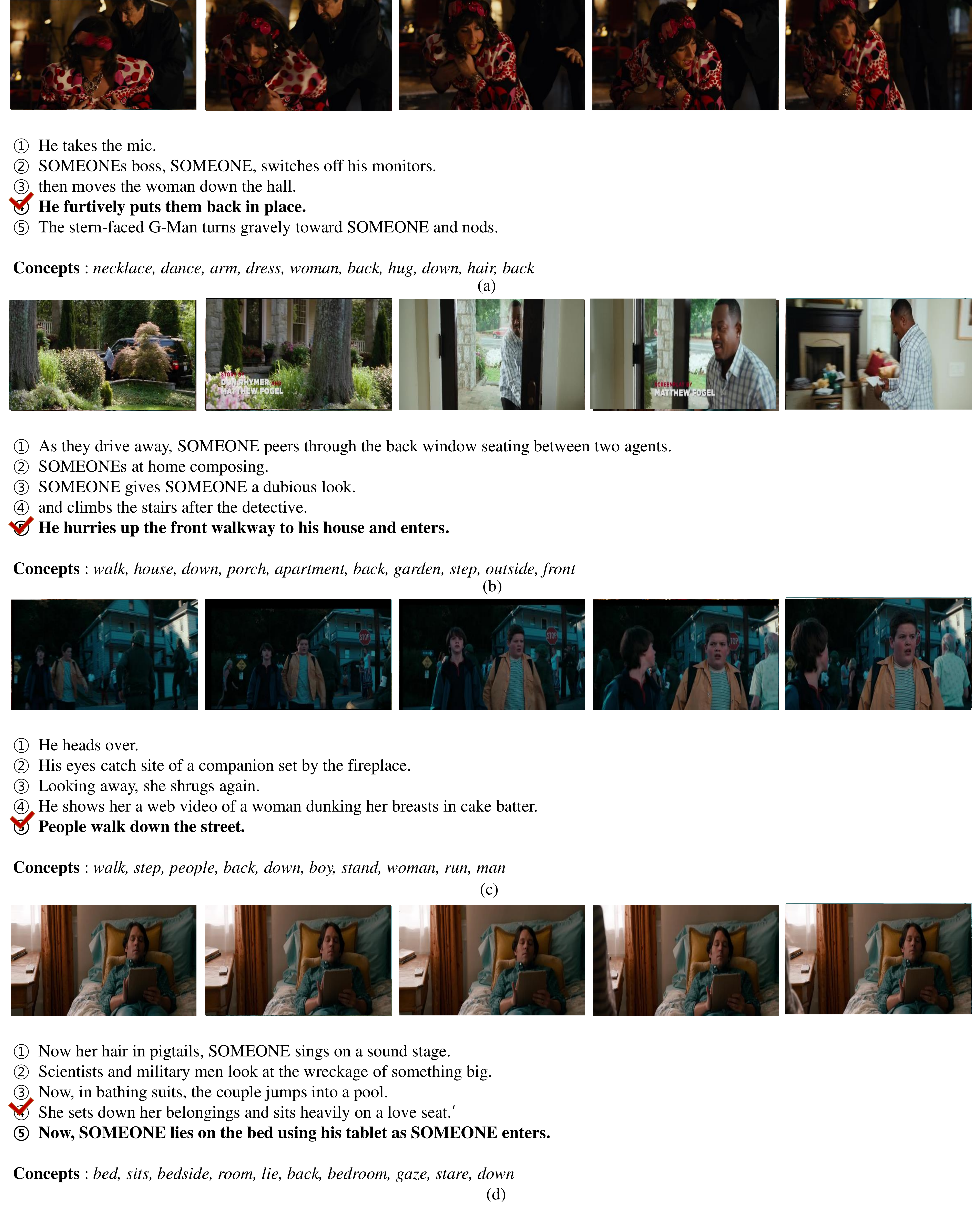}
\caption{Examples of multiple-choice test.
    The groundtruth answer is in bold, and the output of our model is marked with a red checkbox.
    (a)-(c) are positive examples, and (d) is a near-miss or wrong example.
}
\label{fig:example_mc1}
\vspace{-5pt}
\end{figure*}

\begin{figure*}[p]
\centering
\includegraphics[width=0.95\textwidth,trim=0cm 0cm 0.8cm 0cm]{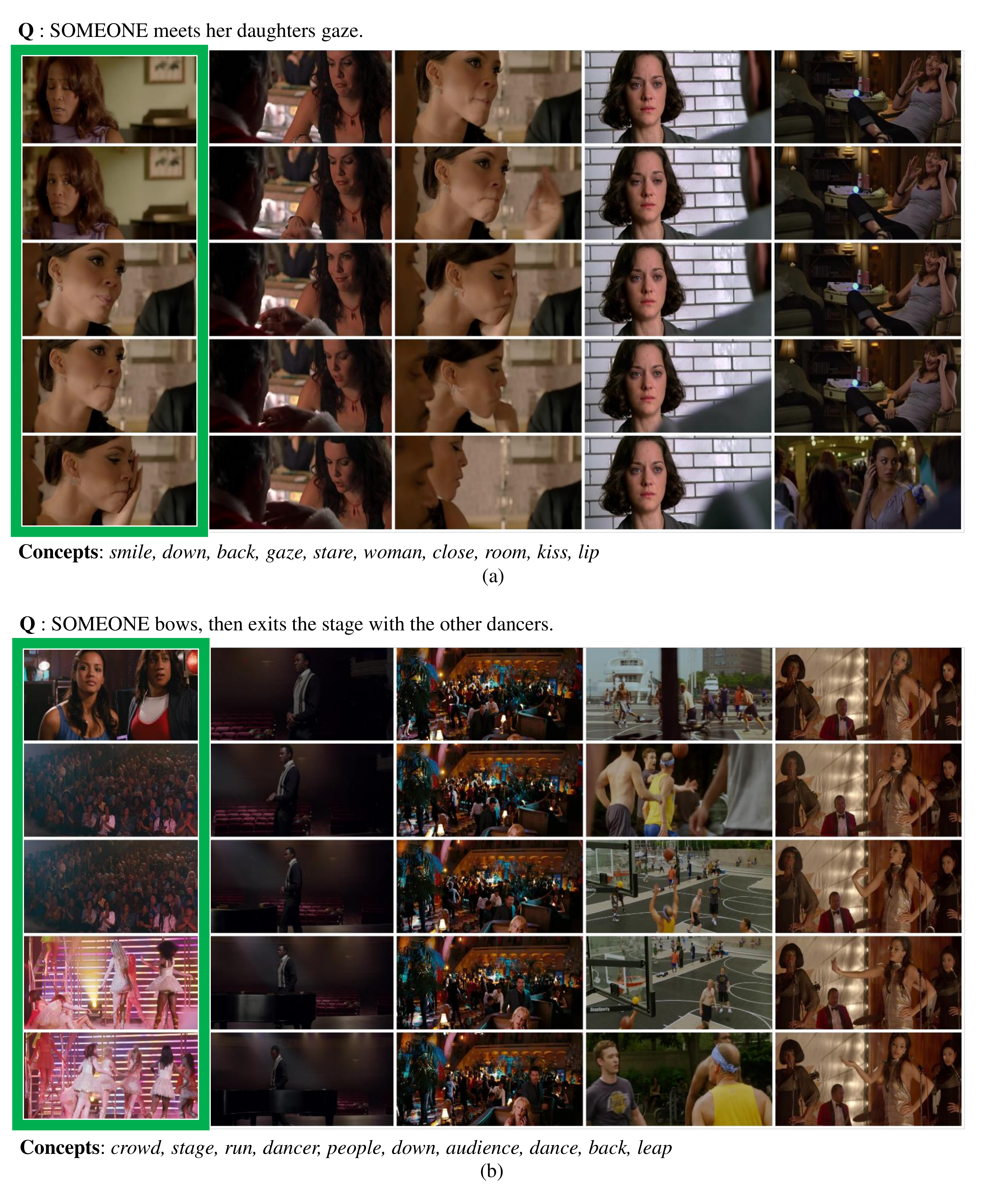}
\caption{Positive examples of movie retrieval. From left to right,
    we show the 1st-5th retrieved movie clips from natural language sentence.
    The groundtruth movie clip is shown in the green box.
}
\label{fig:example_ret1}
\vspace{-5pt}
\end{figure*}

\begin{figure*}[p]
\centering
\includegraphics[width=0.95\textwidth,trim=0cm 0cm 0.8cm 0cm]{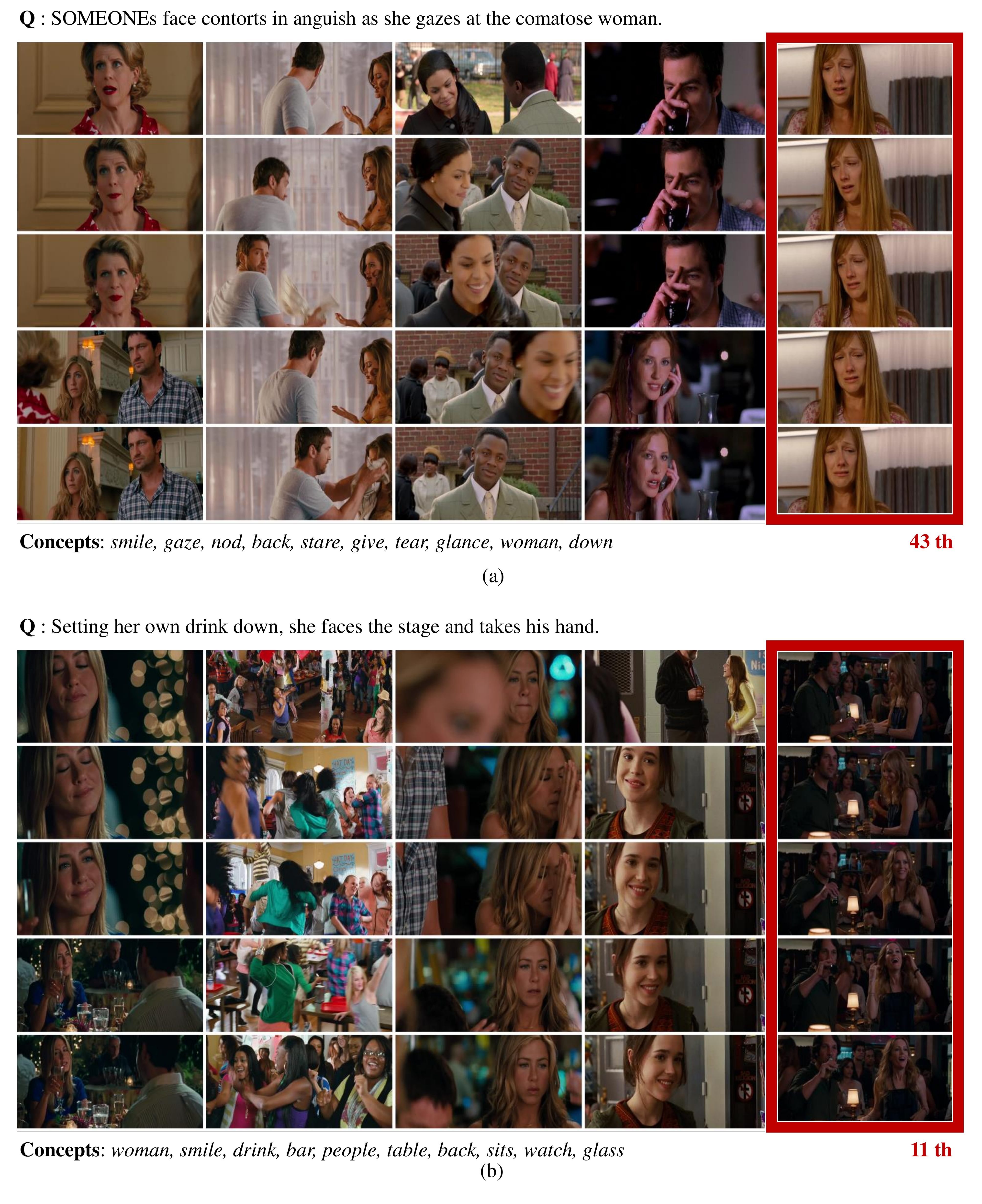}
\caption{Negative examples of movie retrieval.
    The first 4 columns represent the 1st-4th retrieved movie clips,
    and the last one is the groundtruth movie clip (in the red box).
    We also show the retrieved rank of the groundtruth.
}
\label{fig:example_ret3}
\vspace{-5pt}
\end{figure*}

In the following,
we present more examples of movie description results in
Figure \ref{fig:example_moviecap1}.
Additional examples of the fill-in-the-blank task follows
in Figure \ref{fig:example_fib1},
and more examples of the multi-choice test are given in Figure \ref{fig:example_mc1}.
Finally, we present examples of the movie retrieval task in Figure \ref{fig:example_ret1}--\ref{fig:example_ret3}.
We also show each model's output and the detected concept words correspondingly. %

\FloatBarrier


\FloatBarrier

\end{document}